\documentclass[journal=jcisd8,manuscript=article]{achemso}
\SectionNumbersOn
\usepackage[version=3]{mhchem} 

\usepackage{amssymb}
\usepackage{multirow}
\usepackage{comment}

\newcommand{\QED}{\mbox{QED}}
\def\imagetop#1{\vtop{\null\hbox{#1}}}
\author{Zhenpeng Zhou}
\affiliation{Department of Chemistry, Stanford University}
\altaffiliation{Work done during an internship at Google AI Applied Science}
\email{zhenpeng@stanford.edu}
\author{Steven Kearnes}
\affiliation{Google AI Applied Science}
\email{kearnes@google.com}
\author{Li Li}
\affiliation{Google AI Applied Science}
\email{leeley@google.com}
\author{Richard N. Zare}
\affiliation{Department of Chemistry, Stanford University}
\email{rnz@stanford.edu}
\author{Patrick Riley}
\affiliation{Google AI Applied Science}
\email{pfr@google.com}

\title[An \textsf{achemso} demo]
  {Optimization of Molecules via Deep Reinforcement Learning}

\begin{document}





\begin{abstract}
We present a framework, which we call Molecule Deep $Q$-Networks (MolDQN), for molecule optimization by combining 
domain
knowledge of chemistry and state-of-the-art reinforcement learning
techniques (double $Q$-learning and randomized value functions). 
We directly define
modifications on molecules, thereby ensuring 100\% chemical validity.
Further, we operate without pre-training on any dataset 
to avoid possible bias from the choice of that set. 
Inspired by problems faced during medicinal chemistry lead optimization, we extend our model with 
multi-objective reinforcement learning, which maximizes
drug-likeness while maintaining similarity to the original molecule. 
We further show the path through chemical space to achieve optimization
for a molecule to understand how the model works.

\end{abstract}

\section{Introduction}


One fundamental goal in chemistry is to design new molecules with specific 
desired properties. This is especially important in material design 
or drug screening. Currently, this process is expensive in terms of 
time and cost: It can take years and cost millions of dollars to find 
a new drug.\cite{hughes2011principles} The goal of this study
is to partially automate this process through 
reinforcement learning.

To appreciate our approach, it is necessary to review briefly the previous
works that employed machine learning in molecule design.
One prevalent strategy is to build a generative model, which 
maps a point in a high-dimensional latent space to a molecule,
and perform search or optimization in the latent space to find 
new molecules. \citet{gomez2018automatic},
\citet{blaschke2018application},
\citet{segler2017generating}, \citet{lim2018molecular},
and \citet{putin2018adversarial} utilized 
 strings as molecule 
representations to build a generator of SMILES\cite{weininger1988smiles} 
strings, which is a linear string notation
to describe molecular structures.
One of the most challenging goals in this design is to ensure the 
chemical validity of the generated molecules.
\citet{kusner2017grammar} and \citet{dai2018syntax}
added grammar constraints
to SMILES strings to improve the chemical validity of the generated molecules.
Researchers have also built models on graph representations of molecules,
which regards atoms as nodes and bonds as edges in an undirected graph. \citet{li2018learning} and \citet{li2018multi} described
molecule generators that create graphs in a step-wise manner. \citet{de2018molgan} introduced MolGAN for generating small molecular graphs.
\citet{jin2018junction} designed a two-step generation process
in which a tree is first constructed to represent the molecular scaffold and then
expanded to a molecule. Although almost perfect on generating 
valid molecules, these autoencoder-based models usually need to address the
problem of optimization. Most published work uses a separate Gaussian process model
on the latent space for optimization. However, because the latent space is often high dimensional and the objective functions defined on the latent space is usually non-convex, molecule property optimization on 
the latent space can be difficult. 

Another strategy is based on reinforcement learning, which
is a sub-field of artificial intelligence. Reinforcement learning
studies the way to make decisions to achieve the highest reward.
\citet{olivecrona2017molecular}, \citet{guimaraes2017objective}, \citet{putin2018reinforced}, 
and \citet{popova2018deep}
applied reinforcement learning techniques on top of a string
generator to generate the SMILES strings of molecules. They
successfully generated molecules with given desirable properties, but
struggled with chemical validity. Recently, \citet{you2018graph} proposed a graph 
convolutional policy network (GCPN) for generating graph 
representations of molecules with deep reinforcement learning, achieving 100\% 
validity. However, all these methods require pre-training on
a specific dataset. While pre-training makes it easier to generate molecules
similar to the given training set, the exploration ability is limited
by the biases present in the training data.

Here we introduce a new design for molecule optimization by combining chemistry
domain knowledge and reinforcement learning, which we call Molecule Deep $Q$-Networks (MolDQN). We formulate the 
modification of a molecule as a Markov 
decision process (MDP).  \cite{bellman1957markovian}
By only allowing chemically valid actions, we ensure that all the molecules 
generated are valid. 
We then employ the deep reinforcement learning
technique of Deep $Q$-Networks (DQN) \cite{mnih2015human}
to solve this MDP, using the desired properties
as rewards. Instead of pre-training on a dataset, our
model learns from scratch. 
Additionally, with the introduction of multi-objective deep reinforcement
learning, our model is capable of performing 
multi-objective optimization.

Our contribution differs from previous work in three critical aspects:
\begin{enumerate}
    \item All the works presented above use 
    policy gradient methods,
    while ours is based on value function learning. Although policy gradient methods are applicable
    to a wider range of problems, they suffer from high variance when estimating 
    the gradient.\cite{gu2016q} In comparison, in applications where value function learning works, it is usually more 
    stable and sample efficient.\cite{mnih2015human}
    \item Most, if not all, of the current algorithms rely on
    pre-training on some datasets. Although expert pre-training may lead
    to lower variance, this approach limits the search space and
    may miss the molecules which are not in the dataset. 
    In contrast, our method starts from scratch
    and learns from its own experience, which can lead to
    better performance, i.e., discovering molecules with 
    better properties.
    \item Our model is designed for multi-objective reinforcement learning, 
    allowing users to decide the relative importance of each objective. See \ref{multiobjective} for more detail.
    
\end{enumerate}

\section{Methods}

\subsection{Molecule Modification as a Markov Decision Process}\label{sec:mdp}

Intuitively, the modification or optimization of a molecule can be done in a step-wise
fashion, where each step belongs to one of the following three categories: (1) atom
addition, (2) bond addition, and (3) bond removal. The molecule
generated is only dependent on the molecule being changed and the 
modification made. Therefore, the process of molecule 
optimization can be formulated as a Markov decision
process (MDP).
We have several key differences from previous work that employed MDP for molecule modification.~\cite{you2018graph}
\begin{itemize}
    \item We add an explicit limit on the number of steps. This allows us to easily control how far away from a starting molecule we can go. In vast chemical space, this is a very natural way to control the diversity of molecules produced. 
    \item We do not allow chemically invalid actions (violations of valence constraints). These actions are removed from the action space entirely and are not even considered by our model.
    \item We allow atoms/bonds to be removed as well as added.
\end{itemize}

Formally, we have MDP$(\mathcal{S}, \mathcal{A}, \{P_{sa}\}, 
\mathcal{R})$, where we define each term in what follows:
\begin{itemize}
\item $\mathcal{S}$ denotes the state space, in which each state $s \in \mathcal{S}$ is
a tuple of $(m, t)$. Here $m$ is a valid molecule 
and $t$ is the 
number of steps taken. For the initial state, the molecule $m$ can 
be a specific molecule or nothing, and $t=0$.  We limit the maximum 
number of steps $T$ that can be taken in this MDP. In
other words, the set of terminal states is defined as 
$\{s = (m, t) | t=T\}$, which consists of the states whose step number reaches its maximum value. 

\item $\mathcal{A}$ denotes the action space, in which each action $a \in \mathcal{A}$ is
a valid modification to a specific molecule $m$. Each modification belongs to one of the
following three categories mentioned before:
  \begin{enumerate}
  \item Atom addition. Firstly, we define the set of $\mathcal{E}$ be the set of
  elements a molecule contains. We then define a valid action
  as adding (1) an atom in $\mathcal{E}$ and (2) a bond (with all valence-allowed bond orders)
  between the added atom and the original molecule wherever possible. For example, with
  the set of elements 
  $\mathcal{E} = \{ \mathrm{C}, \mathrm{O} \}$, the atom addition action
  set of cyclohexane contains the 4 actions shown in Figure~\ref{fg:c6h6_act}a. Note that hydrogens are considered implicitly, and all atom additions are defined as replacements of implicit hydrogens.

  \item Bond addition. A bond addition action is performed between two atoms with free
  valence (not counting implicit hydrogens). If there is no bond between those two atoms, actions between them consist of
  adding a single, double, or triple bond if the valence allows this change. Additional actions \emph{increase} the bond order between those two atoms by one or two. In other words, the 
  transitions include:
      \begin{itemize}
      \item No bond $\rightarrow$ \{Single, Double, Triple\} Bond.
      \item Single bond $\rightarrow$ \{Double, Triple\} Bond.
      \item Double bond $\rightarrow$ \{Triple\} Bond.
      \end{itemize}
  
  To generate molecules that are chemically more reasonable, we include several heuristics that incorporate chemistry domain
  knowledge.
  First, in order to prevent generating
  molecules with high strain, we do not
  allow bond formation between atoms that are in rings.
  In addition, we added an option that only allows formation 
  of rings with a specific number of atoms. In the experiments we have done, 
  only rings with 3 to 6 atoms are allowed in consideration
  of the most common ring sizes. As an example, Figure~\ref{fg:c6h6_act}b shows the allowed bond addition actions for cyclohexane. 
  \item Bond removal. We define the valid bond removal action set as the actions
  that decrease the bond order of an existing bond. The transitions include:
      \begin{itemize}
      \item Triple bond $\rightarrow$ \{Double, Single, No\} Bond.
      \item Double bond $\rightarrow$ \{Single, No\} Bond.
      \item Single bond $\rightarrow$ \{No\} Bond.
      \end{itemize}
  Note that bonds are only completely removed if the resulting
  molecule has zero or one disconnected atom (and in the latter case, the disconnected
  atom is removed as well). Therefore, no molecules having disconnected parts are
  created in this step.
  \end{enumerate}
    \begin{figure}
      \begin{tabular}{ccc}
    (a) Atom addition & (b) Bond addition & (c) Bond removal\\
    \includegraphics[]{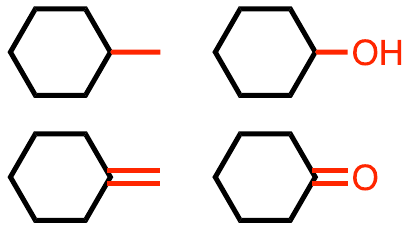} & \includegraphics[]{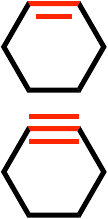} 
      & \multicolumn{1}{c}{\includegraphics[]{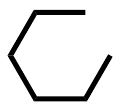}} 
    \end{tabular}
    \caption{Valid actions on the state of cyclohexane. Modifications are shown in red.
    Invalid bond additions which violate the heuristics 
    explained in Section~\ref{sec:mdp} are not shown.}
    \label{fg:c6h6_act}
  \end{figure}
  In our design choice, we do not break an aromatic bond. However, it is still possible to break aromaticity. (See the third molecule in Figure~\ref{fg:multi_obj_scatter}, $w=0.4$; the removal of the extracyclic double bond from the original molecule breaks aromaticity.) Besides, an aromatic system can still be
  created in a stepwise way by adding single and double bonds alternatively, and the resulting system will be perceived as aromatic by the RDKit SMILES parser.
  We also include ``no modification'' as an action, which allows the molecule to remain unchanged before reaching the step limitation~$T$.

\item $\{P_{sa}\}$ denotes the state transition probability. Here we define the state
transition to
be deterministic. For example, if we modify a molecule by adding a single bond, the next
state we reach will be the new molecule adding the bond, with a probability of 1.

\item $\mathcal{R}$ denotes the reward function of state $(m, t)$. In material design or
lead optimization, the reward is often a property of the molecule $m$. 
In our design, a reward is given not just at the terminal states, but
at each step, which empirically  produces better learning performance (see Figure~S3). To ensure that the final state is rewarded most heavily, we discount the value of the rewards at a state with time $t$ by  $\gamma^{T - t}$ (where we typically used $\gamma=0.9$). Note that the definition of discount factor is different from the usual way. In future discussions of reward $r_t$, this discount factor is implicitly included for simplicity.

\end{itemize}

\paragraph{Implementation details.}
We implemented the state transition of a molecule with the available software framework of RDKit.\cite{rdkit}. The properties of molecules are calculated
with tools provided by RDKit.

\subsection{Reinforcement Learning}

Reinforcement Learning is an area of machine learning concerning how the 
\emph{decision makers} (or \emph{agents}) ought to take a series of actions in a prescribed
\emph{environment} so as to maximize a notion of cumulative reward, especially
when a model of the environment is not available. Here, the environment is
the molecule modification MDP we defined above, and our goal is to find a
policy $\pi$ which selects an action for each state that can maximize the 
future rewards.

Intuitively, we are trying to fit a function $Q(s, a)$ that predicts
the future rewards of taking an action $a$ on state $s$. A decision
is made by choosing the action $a$ that maximizes the $Q$ function, which
leads to larger future rewards. 

Mathematically, for a policy $\pi$, 
we can define the value of an action $a$ on a state $s$
to be 
\[
Q^\pi(s, a) = Q^\pi(m, t, a) = \mathbb{E}_{\pi}\left[\sum_{n=t}^T  r_n \right]
\]
where $\mathbb{E}_{\pi}$ denotes taking an expectation with respect to $\pi$,
and $r_n$ denotes the reward at step $n$. This action-value
function calculates the future rewards of taking action
$a$ on state $s$, and subsequent actions decided by policy $\pi$. We can therefore
define the optimal policy $\pi^*(s)= \arg\max_a Q^{\pi^*}(s, a)$. 

In our case, however, we have both a deterministic MDP and an accurate model of the environment. 
Therefore, we chose to approximate the value function $V(s)=\max_a Q(s, a)$ and we calculate the $Q$ function for an action $a$ moving from state $s$ to $s'$ as $Q(s, a) = \mathcal{R}(s') + V(s')$

Under the setting that the maximum number of steps is limited, the MDP is 
time-dependent, and the optimal policy will be time-dependent as well. Naturally,
if there are many steps left, we can risk pursuing later but larger
rewards, while if only a few steps remain, we should focus on rewards that can be obtained sooner.

We adopt a deep $Q$-learning \cite{mnih2015human}
algorithm to find an estimate
of the $Q$ function. We refer to a neural network function
approximator as the parameterized $Q$-value function $Q(s, a; \theta)$, where $\theta$ is the parameter.
This approximator can be trained by minimizing the loss function of 
\[
l(\theta) = \mathbb{E}\left[ f_l\left(y_t  - Q(s_t, a_t; \theta)\right) \right]
\]
where $y_t = r_t + \max_a Q(s_{t+1}, a; \theta)$ is the target
value, and $f_l$ is a loss function. In our case, we use the Huber loss\cite{boyd2004convex} as a loss function.

\begin{equation*}
f_l(x)=
\begin{cases}
  \frac{1}{2} x^2 & \text{if}\ |x|<1 \\
  |x| - \frac{1}{2} & \text{otherwise}
\end{cases}
\end{equation*}

\subsection{Multi-Objective Reinforcement Learning}
\label{morl}
In real-world applications like lead optimization, it is often desired to 
optimize several different properties at the same time. For example, we may
want to optimize the selectivity of a drug while keeping the solubility in a specific range. Formally, under the multi-objective
reinforcement learning setting, the environment will return a
vector of rewards at each step $t$, with one reward for each objective,
i.e. $\vec{r_t} = [r_{1, t}, \cdots, r_{k, t}]^T\in \mathbb{R}^k$, where $k$ is the number
of objectives.

There exist various goals in multi-objective optimization. The goal may be finding a set of Pareto optimal solutions, or find a single or several solutions that satisfy the preference of a decision maker. Similar to the choice in \citet{guimaraes2017objective}, we adapted the latter one in this paper. Specifically, we implemented the ``scalarized'' reward framework to realize multi-objective optimization, with the introduction of a user defined weight vector $w = [w_1, w_2 \cdots, w_k]^T\in \mathbb{R}^k$, the scalarized reward can be calculated as $$r_{s,t} = w^T\vec{r_t} = \sum_{i=1}^k w_i r_{i, t}$$. The objective of the MDP is then to maximize the cumulative scalarized reward.

\subsection{Exploitation vs. Exploration During Training}
The trade-off between exploitation and exploration presents a dilemma caused by the
uncertainty we face. Given that we do not have a complete knowledge
of the rewards for all the states, if we constantly choose the best
action that is known to produce the highest reward (\emph{exploitation}),
we will never learn anything about
the rewards of the other states. On the other hand, if we always
chose an action at random (\emph{exploration}), we would not receive
as much reward as we could achieve by choosing the best action.

One of the simplest and the most widely used approaches to balance these competing goals is called 
$\varepsilon$-greedy, which selects the predicted best action with probability
$1-\varepsilon$, and a uniformly random action with probability
$\varepsilon$. Without considering the level of uncertainty of the value
function estimate, $\varepsilon$-greedy often wastes exploratory
effort on the states that are known to be inferior. 

To counter this issue, we followed the idea of bootstrapped-DQN from \citet{osband2016deep}
by utilizing randomized
value functions to achieve deep exploration. We built $H$ independent $Q$-functions $\{Q^{(i)}|i=1,\cdots,H\}$ (actually, a multi-task neural network with a separate head for each $Q^{(i)}$; see Section~\ref{dqn_details}), each of them being
trained on a different subset of the samples. At each episode, we uniformly
choose $i\in \{1, \cdots, H\}$, and use $Q^{(i)}$ for decision making. 
The above approach is combined with $\varepsilon$-greedy as our policy.
During training, we annealed $\varepsilon$
from 1 to 0.01 in a piecewise linear way.

\subsection{Deep $Q$-Learning Implementation Details}
\label{dqn_details}
We implemented the deep $Q$-learning model described by \citet{mnih2015human} with
improvements of double $Q$-learning\cite{van2016deep}.
Recall that a state $s$ is a pair of molecule $m$ and time $t$. Unsurprisingly, including $t$ in the model performs better experimentally (see Figure~S4).

We used a deep neural network to approximate the 
$Q$-function. The input molecule is converted to a vector form called its Morgan fingerprint\cite{rogers2010extended} with 
radius of 3 and length of 2048, and the number of steps remaining in the episode was concatenated to the vector.  A four-layer
fully-connected network with hidden sizes of [1024, 512, 128, 32] and ReLU activations is used as the network architecture. Its output dimension is the number $H$ (see above; for computational efficiency, we implemented these $H$ different models as multiple outputs on top of shared network layers). In most experiments, we limited the maximum
number of steps per episode to 40, given that most drug molecules have less
than 40 atoms (the exception is for the experiments in Section~\ref{single_prop_opt}, where we limit the max number of steps to be 38 for logP optimization to match \citet{you2018graph}, and Section~\ref{constrained_opt}, where the limit is 20.). We trained the model for 5,000
episodes with the Adam optimizer\cite{kingma2014adam} with a learning rate of 0.0001. We used $\varepsilon$-greedy together
with randomized value functions as a exploration policy, and, as mentioned before, we annealed $\varepsilon$
from 1 to 0.01 in a piecewise linear way. The discount factor $\gamma$ (as defined in Section~\ref{sec:mdp}) was set to 0.9.

\section{Results and Discussion}

In these tasks, we demonstrated the effectiveness of our framework on
optimizing a molecule to achieve desired properties. We compared MolDQN with the following state-of-the-art models:
\begin{itemize}
\item Junction Tree Variational Autoencoder (JT-VAE) \cite{jin2018junction} is a deep
generative model that maps molecules to a high-dimensional 
latent space and performs sampling or optimization in the 
latent space to generate molecules.
\item Objective-Reinforced Generative Adversarial Networks (ORGAN)
\cite{guimaraes2017objective}
is a reinforcement learning based molecule generation algorithm
that uses SMILES strings for input and output.
\item Graph Convolutional Policy Network (GCPN)
\cite{you2018graph} is another
reinforcement learning based algorithm that operates on a graph representation
of molecules in combination with a MDP.
\end{itemize}

\subsection{Single Property Optimization}
\label{single_prop_opt}
In this task, our goal is to find a molecule that can maximize
one selected property. Similar to the setup in previous 
approaches\cite{you2018graph,jin2018junction}, 
we demonstrated the
property optimization task on two targets: penalized logP and  Quantitative Estimate of Druglikeness 
(QED)\cite{bickerton2012quantifying}. LogP is the logarithm of the partition ratio
of the solute between octanol and water.
Penalized logP\cite{jin2018junction} is the logP minus the synthetic accessibility (SA)
score and the number of long cycles.

\begin{figure}
\begin{tabular}{cc}
    	\includegraphics[width=0.47\textwidth]{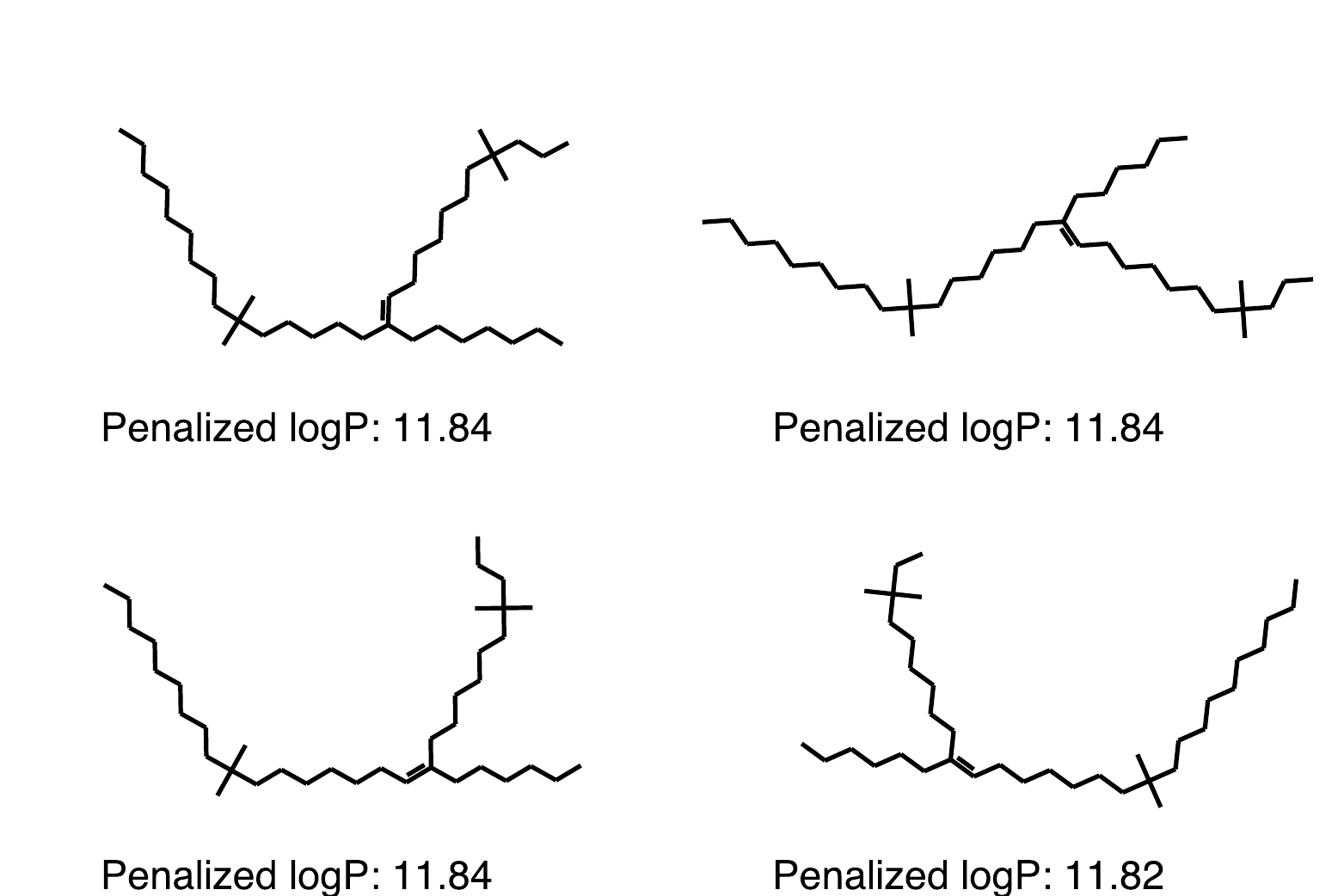} & \includegraphics[width=0.47\textwidth]{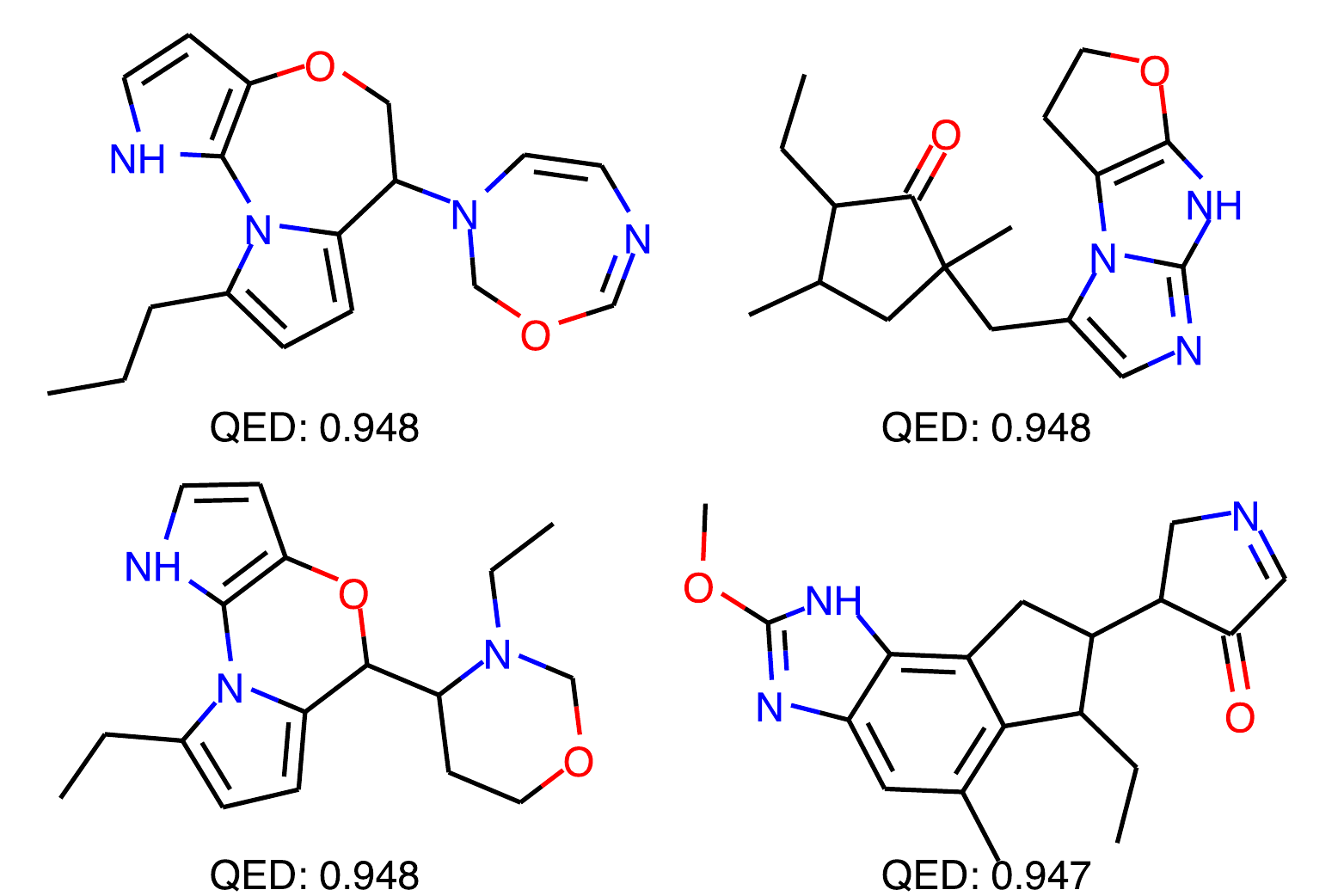} \\
      (a) Optimization of penalized logP & (b) Optimization of QED 
\end{tabular}
  \caption{Sample molecules in the property optimization task. 
  (a) Optimization of penalized logP from MolDQN-bootstrap; note that the generated molecules are obviously not drug-like due to the use of a single-objective reward. (b) Optimization of QED from MolDQN-twosteps.}
  \label{fg:opt_sample}
\end{figure}

In this experiment setup, the reward was set to be the
penalized logP or QED score of the molecule. For logP optimization,
the initial molecule was set to be empty, while for QED optimization,
a two-step optimization was used to improve the result. The first step started with an empty
molecule, and the second step started with the 5 molecules that have the highest
QED values found in step one.
The max number of steps per episode for LogP optimization is set to be 38, in order to allow a direct comparison with GCPN. We will discuss the rationale for this choice in later paragraphs. This number is set to 40 in QED optimization.
We picked the last 100 terminal states in the training process and report the top three property
scores found by each model and the percentage of valid molecules in 
Table~\ref{tb:prop_opt}.
Note that the range of penalized logP is
$(-\infty, \infty)$, while the range of QED is $[0, 1]$. We also visualized the 
best molecules we found in Figure~\ref{fg:opt_sample}. Note that in the optimization
of penalized logP, the generated molecules are obviously not drug-like, which highlights the importance of carefully designing the reward (including using multiple objectives in a medicinal chemistry setting) when using reinforcement learning.

We compared our model to three baselines. ``Random walk'' is a baseline that chooses a random action for each step, ``greedy'' is a baseline that chooses the action that leads to the molecule with the highest reward for each step, and ``$\varepsilon$-greedy'' follows the ``random'' policy with probability $\varepsilon$, and ``greedy'' policy with probability $1-\varepsilon$. Additionally, we compared our model to three published literature models: ORGAN,\cite{guimaraes2017objective} JT-VAE,\cite{jin2018junction} and GCPN\cite{you2018graph}.

With the introduction of bootstrapped DQN, we are able to find molecules with higher QED values compared to naive DQN, demonstrating the exploration efficiency of bootstrapping. However, on the task of maximizing penalized logP, bootstrapped DQN does not provide a significantly better result. This is partly because maximizing logP corresponds to a simple policy: adding carbon atoms wherever possible. This straightforward policy does not require much exploration effort, and can be regarded as a greedy policy (Table~\ref{tb:prop_opt}).

Moreover, our experiments reveal that the task of maximizing logP with no constraints is not a good metric to evaluate the performance of a model. The penalized logP value almost increases linearly with the number of atoms (Figure~S7), therefore it is not fair to compare logP without limiting the number of atoms to be the same. Although the task of optimizing logP can be used to evaluate whether a model can capture the simple domain-specific heuristic, we suggest that maximization should be performed under certain constraints, for example, number of atoms, or similarity. We also suggest that targeting a specific range of logP is also a valid task to evaluate the performance of different models. This task not only avoids the problem of unconstrained optimization, but also represents a real need in typical drug discovery projects.

Compared with GCPN, MolDQN demonstrates better performance on the task of logP, and similar performance on the task of QED. These results
can be partly attributed to
learning from scratch, where the scope is not limited to the 
molecules in a specific dataset.

\begin{table}
  \caption{Top three unique molecule property scores found by each method.}
  \label{tb:prop_opt}
  \begin{tabular}{ccccccccc}
    \hline
    	   & \multicolumn{4}{c}{Penalized logP} & \multicolumn{4}{c}{QED} \\ 
           &  1st  &  2nd  &  3rd  &Validity&  1st  &  2nd  &  3rd  &Validity \\
    \hline
    random walk\textsuperscript{\emph{a}}
    	& -3.99 & -4.31 & -4.37 & 100\% & 0.64 & 0.56 & 0.56 & 100\% \\
    greedy\textsuperscript{\emph{b}}
    	& 11.41 & - & - & 100\% & 0.39 & - & - & 100\% \\
    $\varepsilon$-greedy, $\varepsilon=0.1$\textsuperscript{\emph{b}}
    	& 11.64 & 11.40 & 11.40 & 100\% & 0.914 & 0.910 & 0.906 & 100\% \\
    JT-VAE\textsuperscript{\emph{c}}
    	& 5.30  & 4.93  & 4.49  & 100\%  & 0.925 & 0.911 & 0.910 & 100\% \\
    ORGAN\textsuperscript{\emph{c}}  
    	& 3.63  & 3.49  & 3.44  & 0.4\%  & 0.896 & 0.824 & 0.820 & 2.2\% \\
    GCPN\textsuperscript{\emph{c}}   
    	& 7.98  & 7.85  & 7.80  & 100\%  & 0.948 & 0.947 & 0.946 & 100\% \\
    MolDQN-naive   & 11.51  & 11.51  & 11.50 & 100\%  
           & 0.934  & 0.931  & 0.930 & 100\% \\
    MolDQN-bootstrap   & 11.84  & 11.84  & 11.82 & 100\%  
       & 0.948  & 0.944  & 0.943 & 100\% \\
    MolDQN-twosteps   & -  & -  & - & -  
   & 0.948 & 0.948 & 0.948 & 100\% \\
    \hline
  \end{tabular}
  
    \raggedright
      \vspace{0.1in}
      {\small
      \textsuperscript{\emph{a}} ``random walk'' is a baseline that chooses a random action for each step.
      
      \textsuperscript{\emph{b}} ``greedy'' is a baseline that chooses the action that leads to the molecule with the highest reward for each step. ``$\varepsilon$-greedy'' follows the ``random'' policy with probability $\varepsilon$, and ``greedy'' policy with probability $1-\varepsilon$. In contrast, the $\varepsilon$-greedy MolDQN models choose actions based on predicted $Q$-values rather than rewards.
     
     \textsuperscript{\emph{c}} values are reported in \citet{you2018graph}.
     }
\end{table}

\subsection{Constrained Optimization}
\label{constrained_opt}
We performed molecule optimization under a specific constraint, where
the goal is to find a molecule $m$ that has the largest improvement compared
to the original molecule $m_0$, while maintaining similarity 
$\mbox{SIM}(m, m_0) \geq \delta$
for a threshold $\delta$. Here we defined the similarity as the Tanimoto 
similarity\footnote{The Tanimoto similarity uses the ratio of the 
intersecting set to the union set as the measure of similarity. Represented as a mathematical equation $T(a, b) = \frac{N_c}{N_a + N_b - N_c}$.  $N_a$ and $N_b$ represents the number of attributes in each object $(a,b)$. $N_c$ is the number of attributes in common.} between Morgan fingerprints\cite{rogers2010extended} with radius 2 of
the generated molecule $m$ and the original molecule $m_0$. Following the experiment
in \citet{jin2018junction}, we trained a model in an environment 
whose initial state was randomly set to be one of the 800 molecules in ZINC\cite{irwin2012zinc} dataset
which have the \emph{lowest} penalized logP value, and ran the trained
model on each molecule for one episode. The maximum number of steps per episode was limited to 20 in consideration of computational efficiency. In this task, the reward was designed as follows:
\[
\mathcal{R}(s) = \begin{cases}
		\mbox{logP}(m) - \lambda \times (\delta - \mbox{SIM}(m, m_0))						 & \mbox{if}\hspace{1em} \mbox{SIM}(m, m_0) < \delta \\
		\mbox{logP}(m)  & \mbox{otherwise}
	\end{cases}
\]
where $\lambda$ is the coefficient to balance the similarity and logP.
If the similarity constraint is not satisfied, the reward is penalized by 
the difference between the target and current similarity. In our experiments $\lambda =100$. We report the success rate---the percentage of molecules satisfying the similarity constraint---as validity,
as well as the average improvement on logP in Table~\ref{tb:cst_opt}. Using Welch's $t$-test\cite{welch1947generalization} for $N=800$ molecules, we found that both variants of MolDQN gives a highly statistically significant improvement over GCPN for all values of $\delta$ with $t<-8$. The bootstrap variant also significantly outperforms the naive model (except for $\delta=0.2$) with $t<-3$

\begin{table}
  \caption{Mean and
standard deviation of penalized logP improvement in constraint optimization tasks. $\delta$ is the threshold of the 
  similarity constraint $\mbox{SIM}(m, m_0) \geq \delta$. The  success  rate is the  percentage  of molecules satisfying the similarity constraint.}
  \label{tb:cst_opt}
  \footnotesize
  \begin{tabular}{ccccccccc}
    \hline
    	\multirow{2}*{$\delta$}   &
    	\multicolumn{2}{c}{JT-VAE\textsuperscript{\emph{a}}} &
    	\multicolumn{2}{c}{GCPN\textsuperscript{\emph{a}}} &
    	\multicolumn{2}{c}{MolDQN-naive} &
    	\multicolumn{2}{c}{MolDQN-bootstrap} \\ 
           &  Improvement   &  Success   &  Improvement  &  Success   &  Improvement  &  Success &  Improvement  &  Success\\
    \hline

    0.0    & $1.91\pm2.04$  &  97.5\%    & $4.20\pm1.28$ & 100\% & $6.83\pm1.30$ & 100\% & $7.04\pm1.42$ & 100\%\\
    0.2    & $1.68\pm1.85$  &  97.1\%    & $4.12\pm1.19$ & 100\% & $5.00\pm1.55$ & 100\%  & $5.06\pm1.79$ & 100\%\\
    0.4    & $0.84\pm1.45$  &  83.6\%    & $2.49\pm1.30$ & 100\% & $3.13\pm1.57$ & 100\%  & $3.37\pm1.62$ & 100\%\\
    0.6    & $0.21\pm0.71$  &  46.4\%    & $0.79\pm0.63$ & 100\% & $1.40\pm1.05$ & 100\%  & $1.86\pm1.21$ & 100\%\\
    \hline
  \end{tabular}
  
    \raggedright
    \vspace{0.1in}
    {\small
     \textsuperscript{\emph{a}} values are reported in \citet{you2018graph}.
     }
\end{table}

\subsection{Multi-Objective Optimization}
\label{multiobjective}
In drug design, there is often a minimal structural
basis that a molecule must retain to bind a specific target, referred to as the molecular scaffold. This scaffold is usually defined as a molecule with removal of all side
chain atoms. \cite{garg2012scaffold}
Often the question arises: can we
find a molecule similar to a existing one but having a better performance?
We designed the experiment of maximizing the
QED of a molecule while keeping it similar to a starting molecule. The multi-objective reward of a
molecule $m$ was set to be 
a 2-dimensional vector of $\vec{r} = [\QED(m), \mbox{SIM}(m, m_0)]$, where $\QED(m)$ 
is the QED score
and $\mbox{SIM}(m, m_0)$ is the Tanimoto similarity between the Morgan fingerprints of molecule $m$ and the original molecule $m_0$.

\begin{figure}
\begin{tabular}{ll}
  (a) & (b) \\
  \includegraphics[width=0.5\textwidth]{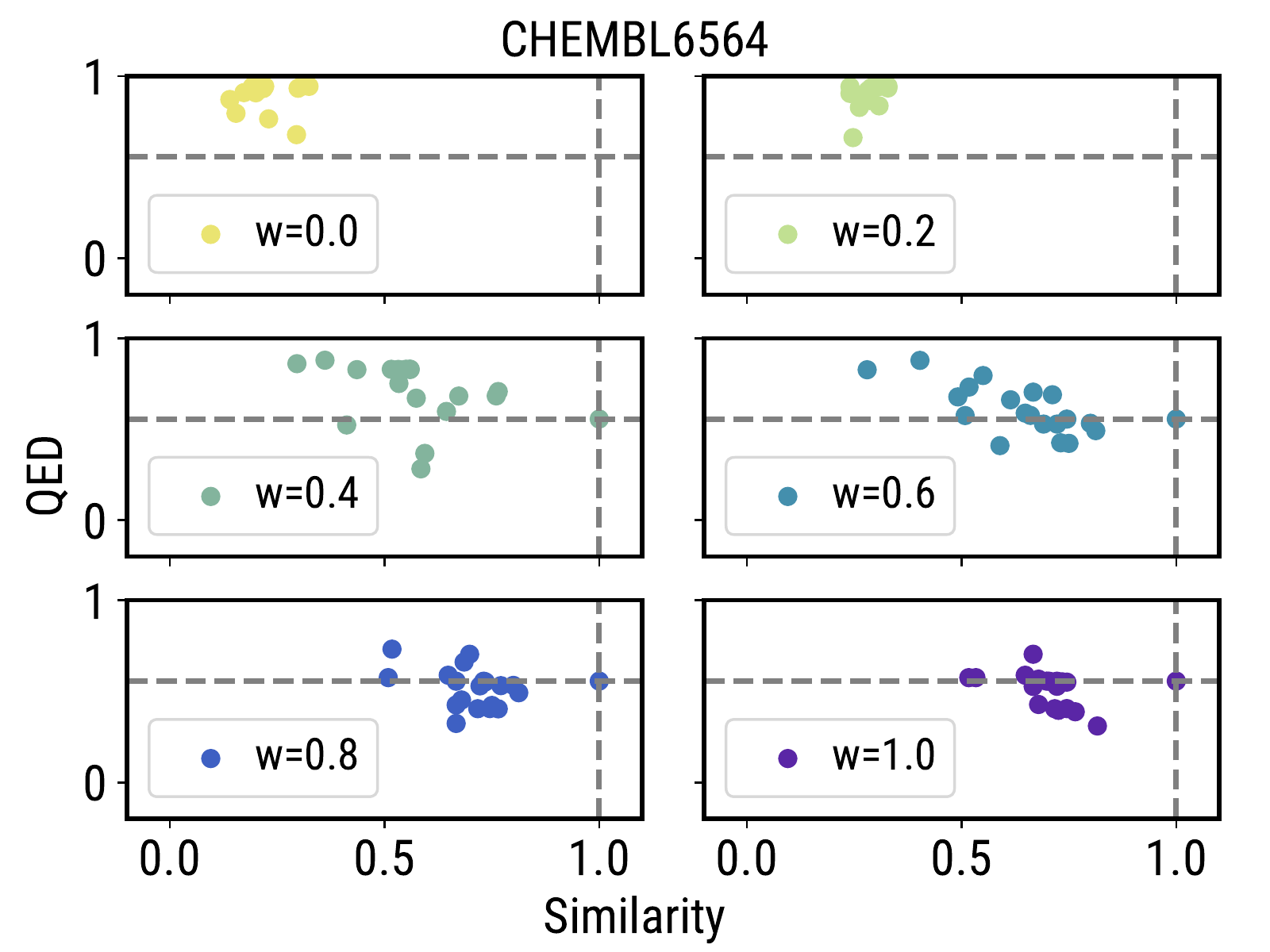} & \includegraphics[width=0.5\textwidth]{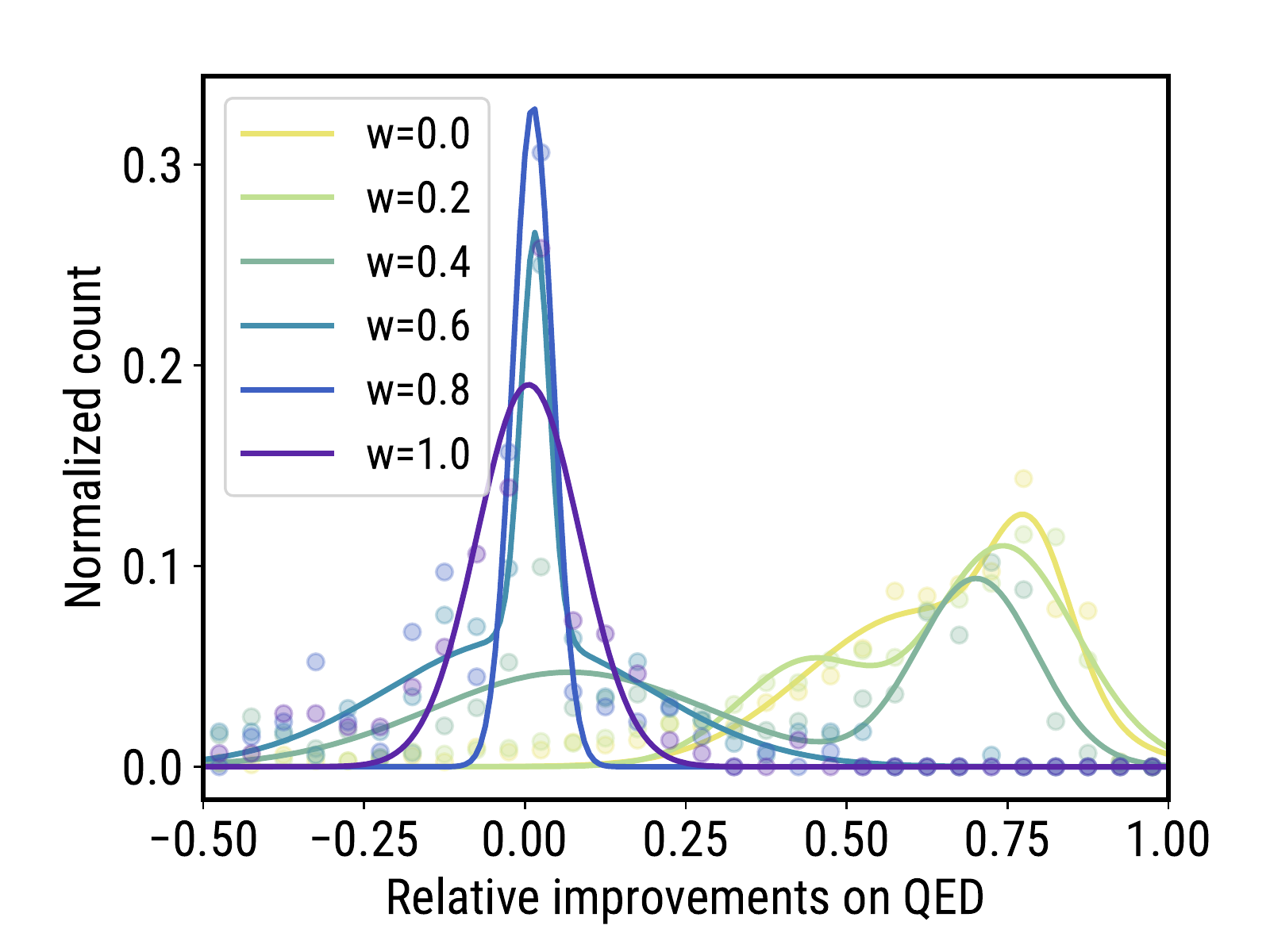} 
\end{tabular}

	\hspace*{-1cm}
	\begin{tabular}{ll}
	(c) & \\
	$w=0.0$ & $w=0.2$ \\
	 \includegraphics[width=0.55\textwidth]{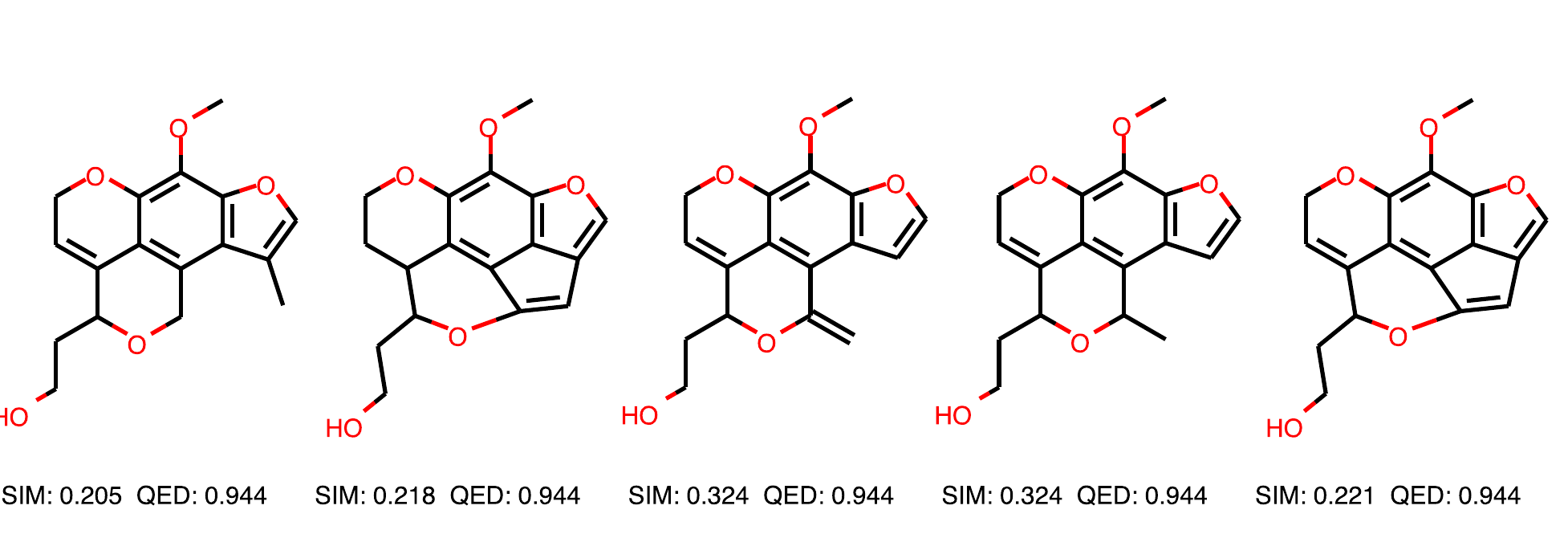} &
	 \includegraphics[width=0.55\textwidth]{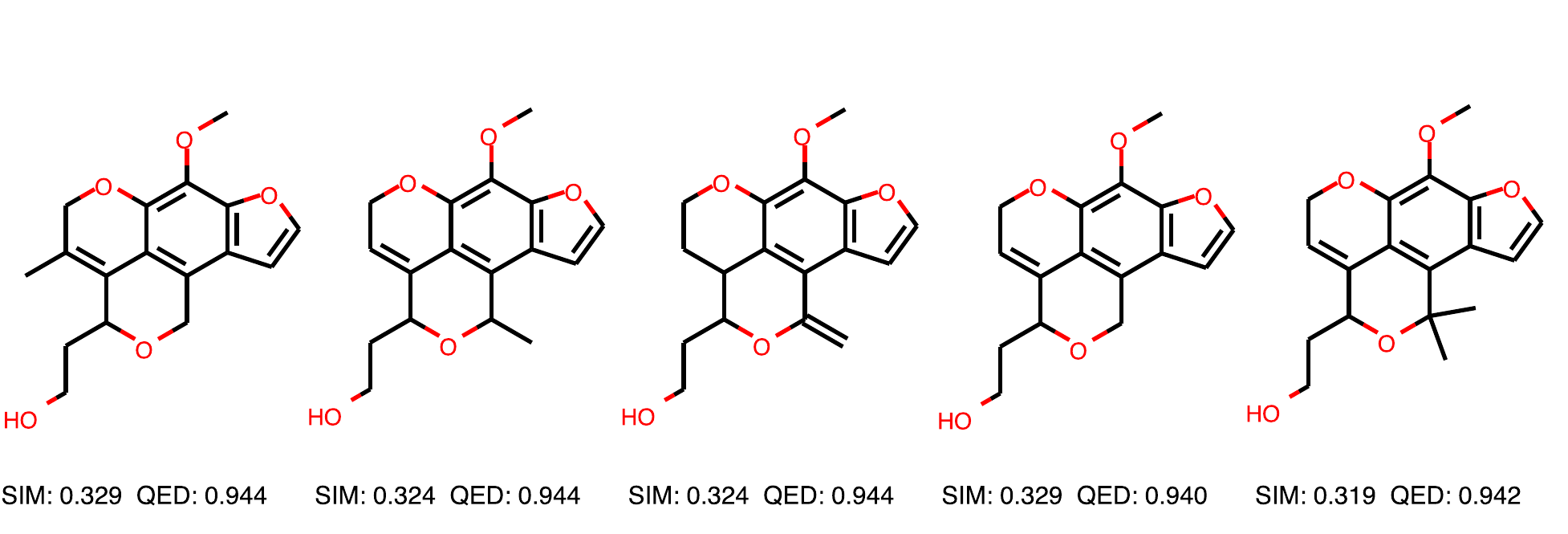} \\
	 $w=0.4$ & $w=0.6$ \\
	\includegraphics[width=0.55\textwidth]{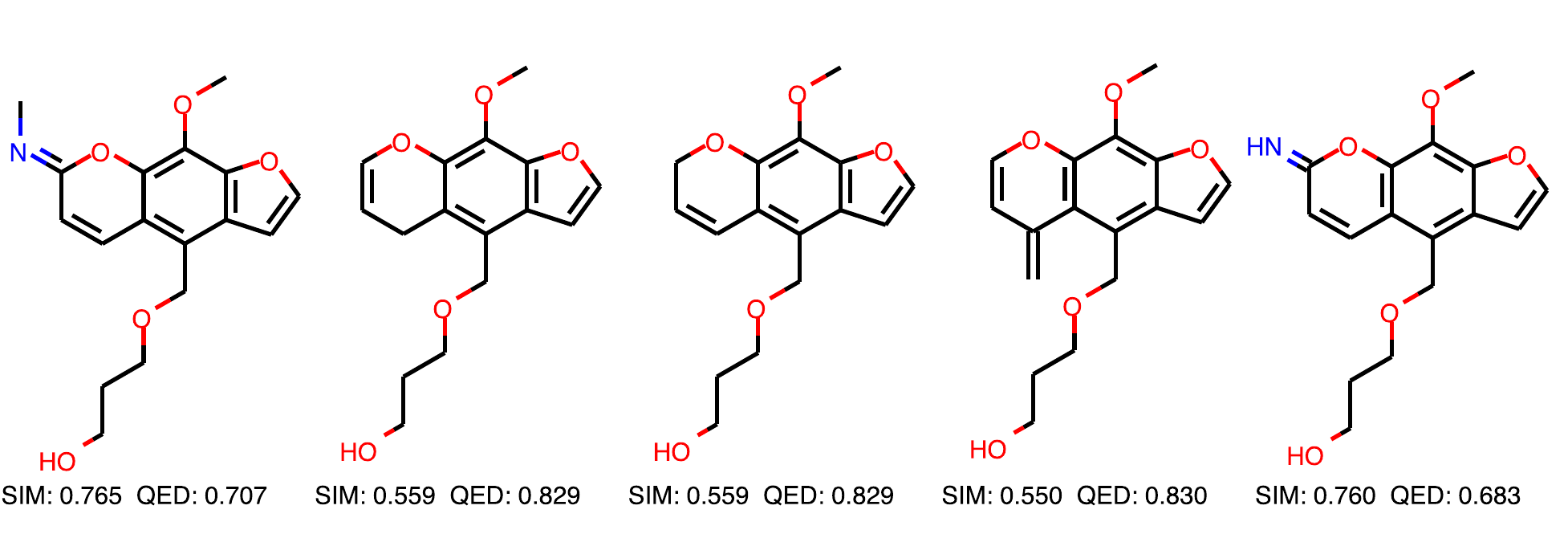} &
	\includegraphics[width=0.55\textwidth]{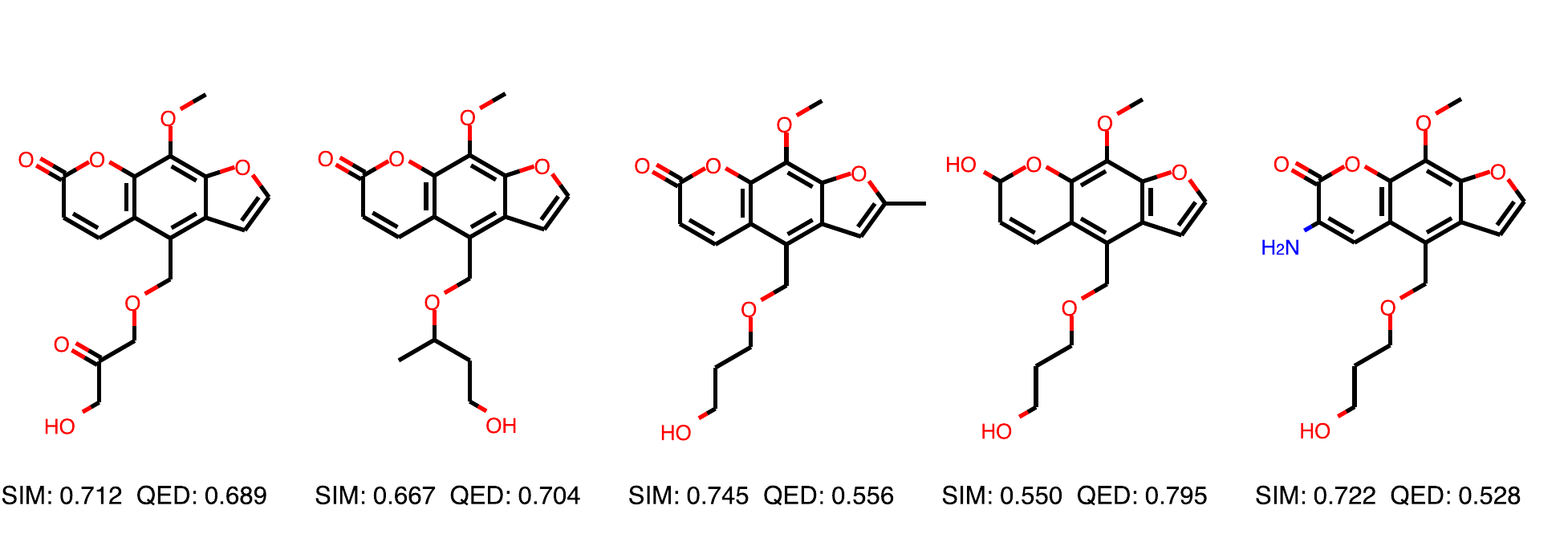} \\
	$w=0.8$ & $w=1.0$ \\
	\includegraphics[width=0.55\textwidth]{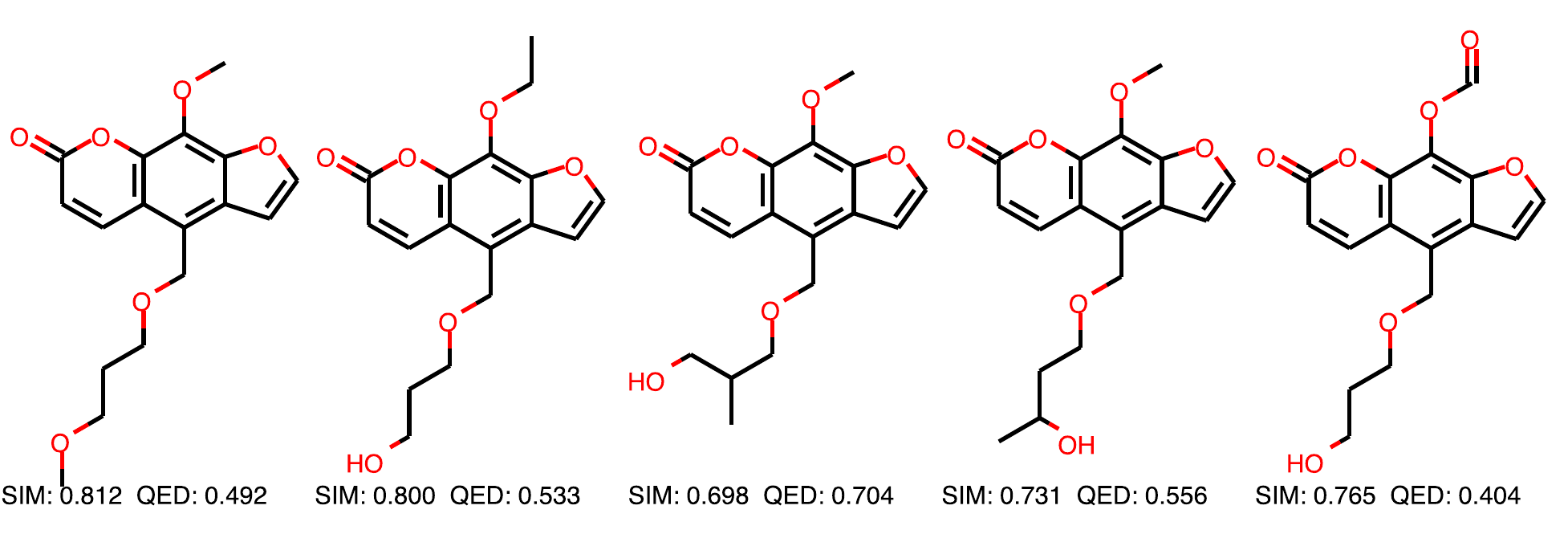} &
	\includegraphics[width=0.55\textwidth]{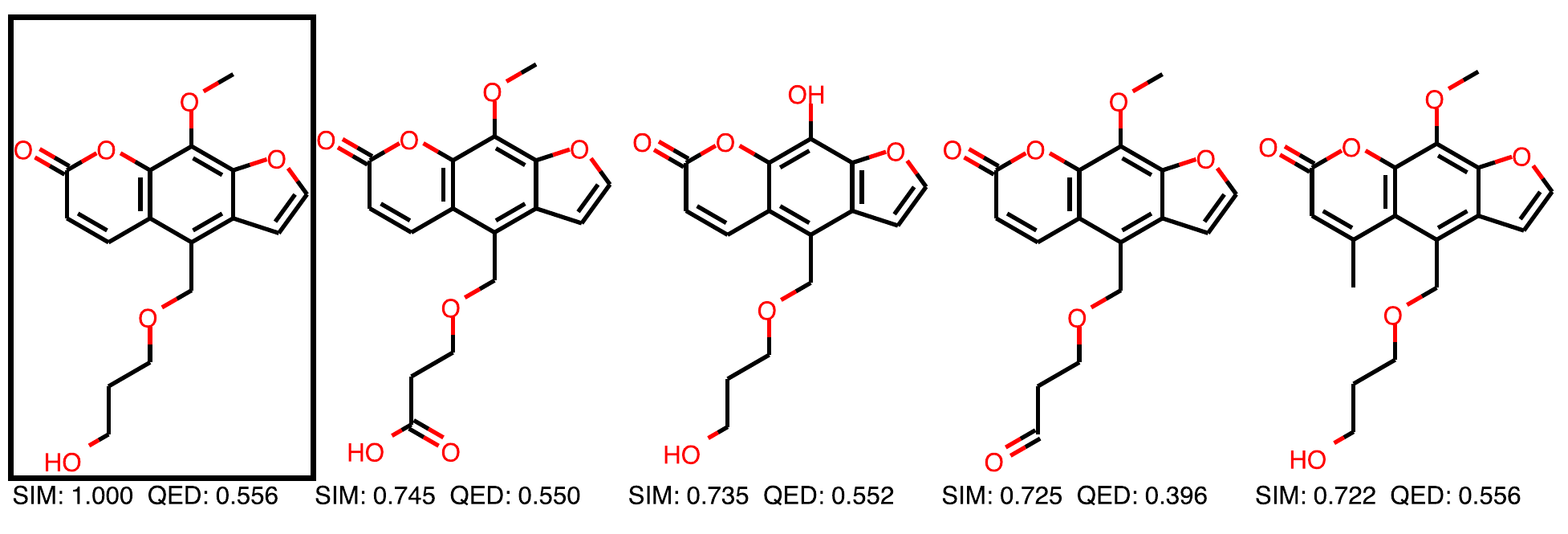} \\
	\end{tabular}

  \caption{(a) The QED and Tanimoto similarity of the molecules optimized under different objective weights. The grey dashed line shows the QED and 
  similarity score of the starting molecule. The legends are transparent, thus it will not cover any point.(b) The empirical 
  distribution of the relative QED improvements in
  20 multi-objective optimization tasks.
  The variable $w$ in legends denotes the weight of the similarity in the multi-objective reward, while the QED score is weighted by ${(1-w)}$, i.e.
$ r = w \times \mbox{SIM}(s) + (1 - w) \times \QED(s)$.
(c) Unique molecules sampled from the multi-objective optimization
  task. The original molecule is boxed.}
  \label{fg:multi_obj_scatter}
\end{figure}

Different weights $w$ can be applied to denote the priorities of these
two objectives. The variable $w$ denotes the weight of similarity
score, while the QED score is balanced by $(1-w)$. This is 
referred to as a ``scalarized'' multi-objective optimization strategy (see Section~\ref{morl}):
\[
\mathcal{R}(s) =w \times \mbox{SIM}(s) + (1 - w) \times \QED(s)
\]
We trained the model with
objective weight of $0.0, 0.2, 0.4, 0.6, 0.8$, and $1.0$, and collected the last 20 unique molecules generated in the training process to plot the properties of molecules on a 2-D space. (i.e., there was no separate evaluation step). Figure~\ref{fg:multi_obj_scatter}a shows the properties
of the optimized molecules under different weights. 
Figure~\ref{fg:multi_obj_scatter}a demonstrates that we can successfully
optimize the QED of a molecule while keeping the optimized molecule similar
to the starting molecule. As the weight applied on similarity
increases, the optimized molecules have higher similarity to the 
starting molecule, and larger fractions of the optimized molecules 
have QED values lower than those of the starting molecules. 
The same experiment was repeated for 20 molecules randomly selected
from ChEMBL\cite{gaulton2016chembl} (Figure S1), and the empirical distribution of the relative improvement of
QED was plotted in Figure~\ref{fg:multi_obj_scatter}b, where the relative improvement of molecule $m$ with respect to the original molecule $m_0$ is defined as

\[
\mathrm{imp}_{rel}=\frac{\mathrm{QED}(m) - \mathrm{QED}(m_0)}{1- \mathrm{QED}(m_0)}
\]

Intuitively, the relative improvement is the ratio of the actual improvement to the largest possible improvement in QED. The distribution of absolute QED improvement is shown in Figure S6.

As the weight
on similarity increases, the distribution of QED improvements moves leftwards because higher priority is placed on similarity. Finally, we
visually examined the optimized molecules (Figure~\ref{fg:multi_obj_scatter}c).
The molecules generated under $w >= 0.4$ possessed the same scaffold as the starting
molecule, indicating that the trained model preserves the original scaffold when the similarity weight is high enough.

\subsection{Optimality vs. Diversity}
Related work in this area reports results for two distinct tasks: optimization and generation (or, to avoid ambiguity, \emph{property-directed sampling}). Optimization is the task to find the best molecule with regard to some objectives, whereas property-directed sampling is the task of generating a set of molecules with specific property values or distributions. 

For the results we report in this paper, we note that there is often a trade-off between optimality
and diversity. Without the introduction of randomness, execution of our learned policy will lead to \emph{exactly one} molecule. Alternatively, there are three possible ways to
increase the diversity of the molecules generated:
\begin{enumerate}
    \item Choose one $Q$ function $Q^{(i)}(s, a)$ uniformly for $i$ in $1, \cdots, H$ to make decision in each episode.
    \item Draw an action stochastically with probability proportional to
          the $Q$-function in each step (as in \citet{haarnoja2017reinforcement}).
    \item During evaluation, use non-zero $\varepsilon$ in the $\varepsilon$-greedy
          algorithm (we took this approach in Section~\ref{multiobjective}).
\end{enumerate}

All of these strategies are sub-optimal because the policy is
no longer pursuing the maximum future rewards. In the results above, 
we focused primarily on optimization tasks and leave the 
question of diversity for future work.

We also conducted experiments to illustrate that we are able to find molecules
with properties in specific ranges with 100\% success (Table S1).
In addition, we demonstrated that we can generate molecules that satisfy multiple
target values (Table S2).
However, because we formulated the property targeting to be an optimization task,
it is not fair for us to compare to other generative models that produce diverse distributions of molecules.

\subsection{Visualization and Interpretation}
Users prefer interpretable solutions when they applying methods that construct new molecules.. Here we demonstrated the 
decision making process of MolDQN that maximizes the QED, starting from a specific molecule.

\begin{figure}
\begin{tabular}{ll}
  (a) & (b) \\
  \imagetop{\includegraphics[width=0.5\textwidth]{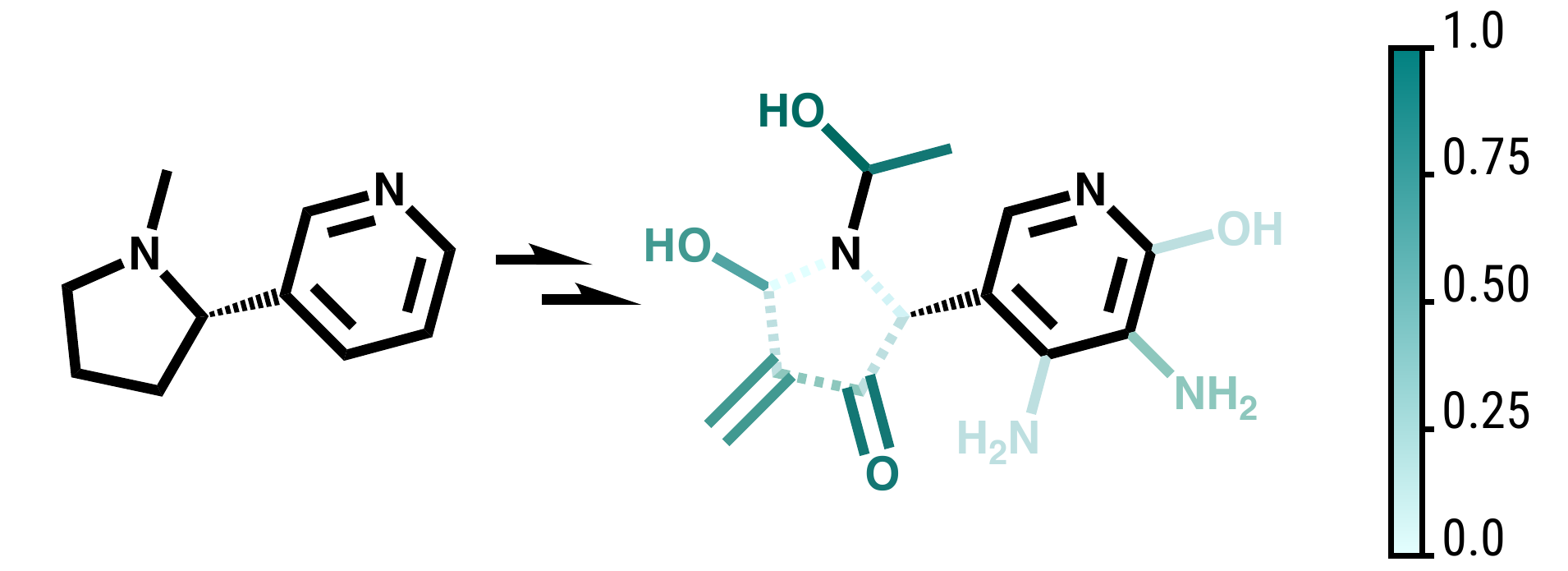}} & \imagetop{\includegraphics[width=0.5\textwidth]{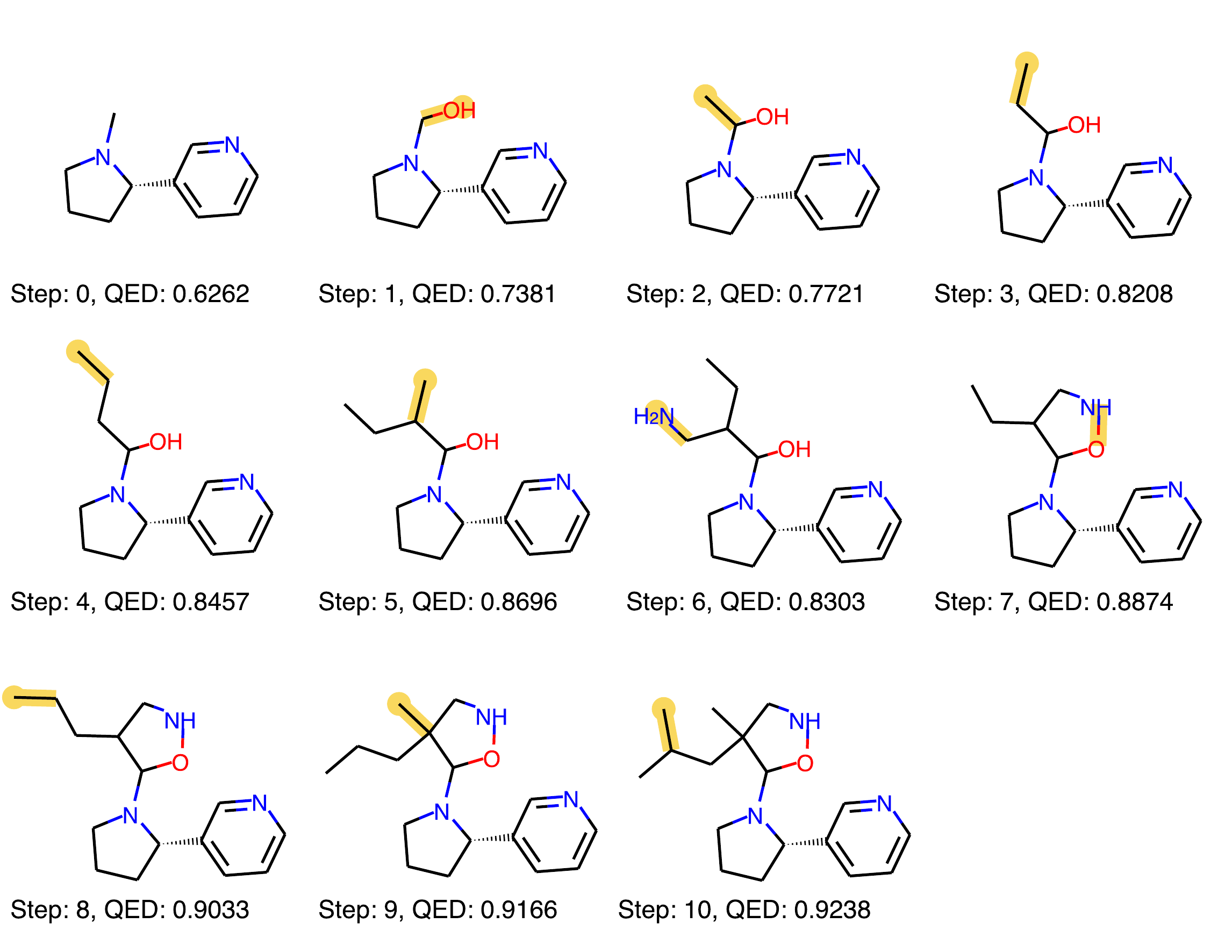}} \\
\end{tabular}

\caption{(a) Visualization of the $Q$-values of selected actions. The full set
of $Q$-values of actions are shown in Figure S2. The original
atoms and bonds are shown in black while modified ones are colored. Dashed
lines denote bond removals. The $Q$-values are rescaled to $[0, 1]$
(b) The steps taken to maximize the QED starting from a molecule. The 
modifications are highlighted in yellow. The QED
values are presented under the modified molecules.}
\label{fg:vis}
\end{figure}

In the first step of decision making, the $Q$-network predicts 
the $Q$-value of each action. Figure~\ref{fg:vis}a shows the predicted $Q$-values
of the chosen actions. The full set
of $Q$-values of for all actions in the first step are shown in Figure S2. We
observe that adding a hydroxyl group
is strongly favored, while breaking the five-member ring structure is disfavored.

Note that the $Q$-value is a measure of future rewards; therefore,
it is possible for the algorithm to choose an action that decreases
the property value in the short term but can reach higher future rewards. Figure~\ref{fg:vis}b shows a sample trajectory of maximizing the QED of a molecule. In this trajectory, step 6 decreases the QED of the molecule, but the QED was improved by 0.297 through the whole trajectory.

\section{Conclusion}
By combining state-of-the-art deep reinforcement learning with
domain knowledge of chemistry, we developed the MolDQN model for molecule optimization. 
We demonstrated that MolDQN reaches equivalent or better performance when compared with several other established algorithms 
in generating molecules with better specified properties. We also presented a way to 
visualize the decision
making process to facilitate learning a strategy for optimizing molecular
design. Future work can be done on applying different $Q$-function approximators (for example MPNN\cite{gilmer2017neural}) and hyperparameter searching. We hope the MolDQN model will assist medicinal and material chemists in molecular design.

As a parting note, it seems obvious to us that the experiments and metrics commonly employed in the literature (including this work) are inadequate for evaluating and comparing generative models in real-world optimization tasks. In particular, logP is a ``broken'' metric that should be discouraged except as a sanity check, and many other commonly used metrics such as QED suffer from boundary effects that limit comparability. Additionally, ``computable'' metrics like QED should be deprioritized in favor of therapeutically relevant properties that can be verified by experiment---this likely requires incorporating predictive models based on experiment into generative decision making, as in \citet{li2018multi}. Even better would be to couple these predictions with experimental validation, as has been done by \citet{merk2018novo} and \citet{putin2018adversarial}. We note that some efforts have been made in addressing generator evaluation\cite{benhenda2017chemgan}, but there remains much work to be done to fairly compare one model to another on meaningful tasks and make these models relevant and effective in prospective drug discovery.

\begin{acknowledgement}

The authors thank Zan Armstrong for her expertise and help in
visualization of the figures. The authors thank David Belanger and John Platt for the internal review and comments. ZZ and RNZ thank the support from the 
National Science Foundation under the Data-Driven Discovery Science 
in Chemistry (D3SC) for EArly concept Grants for Exploratory Research (EAGER)
(Grant CHE-1734082).

\end{acknowledgement}

\begin{suppinfo}

Supporting Information will be available online.

\end{suppinfo}

\section*{Author Contributions}
Z.Z., S.K., L.L., and P.R. conceived the presented idea and performed the computations. P.R. and R.N.Z. supervised the findings of this work. All authors discussed the results and contributed to the final manuscript.
 
\section*{Competing Interests}
The authors declare no competing interests.

\section*{Data Availability}
The ChEMBL\cite{gaulton2016chembl} and ZINC\cite{irwin2012zinc} datasets used in this study are available online. No dataset was generated during the current study. The code is available at \url{https://github.com/google-research/google-research/tree/master/mol_dqn}.

\bibliography{references}

\newpage
\section*{Figure Captions}
\listoffigures

\end{document}


\begin{figure*}
\begin{tabular}{ll}
  \includegraphics[width=0.5\textwidth]{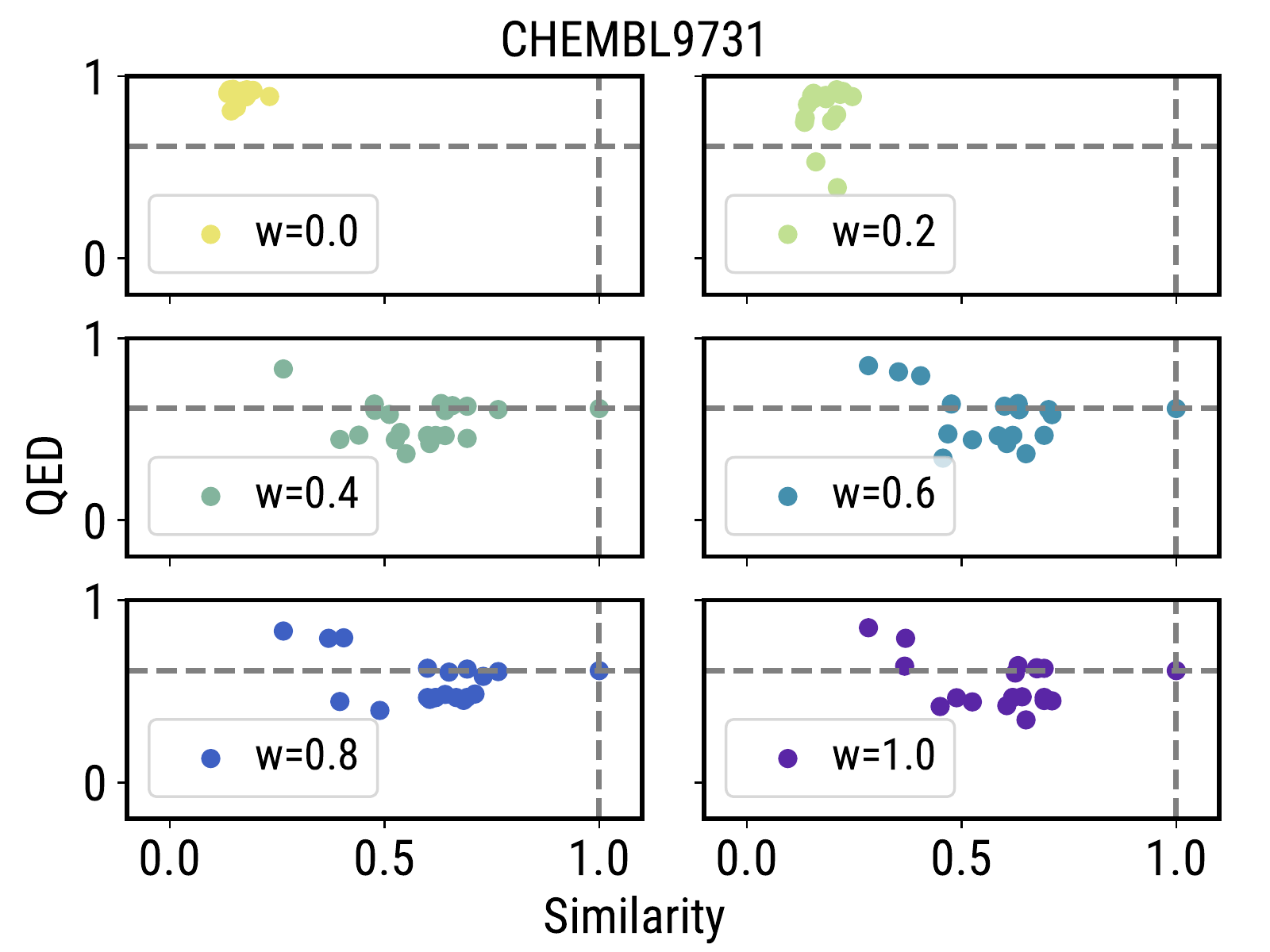} & 
  \includegraphics[width=0.5\textwidth]{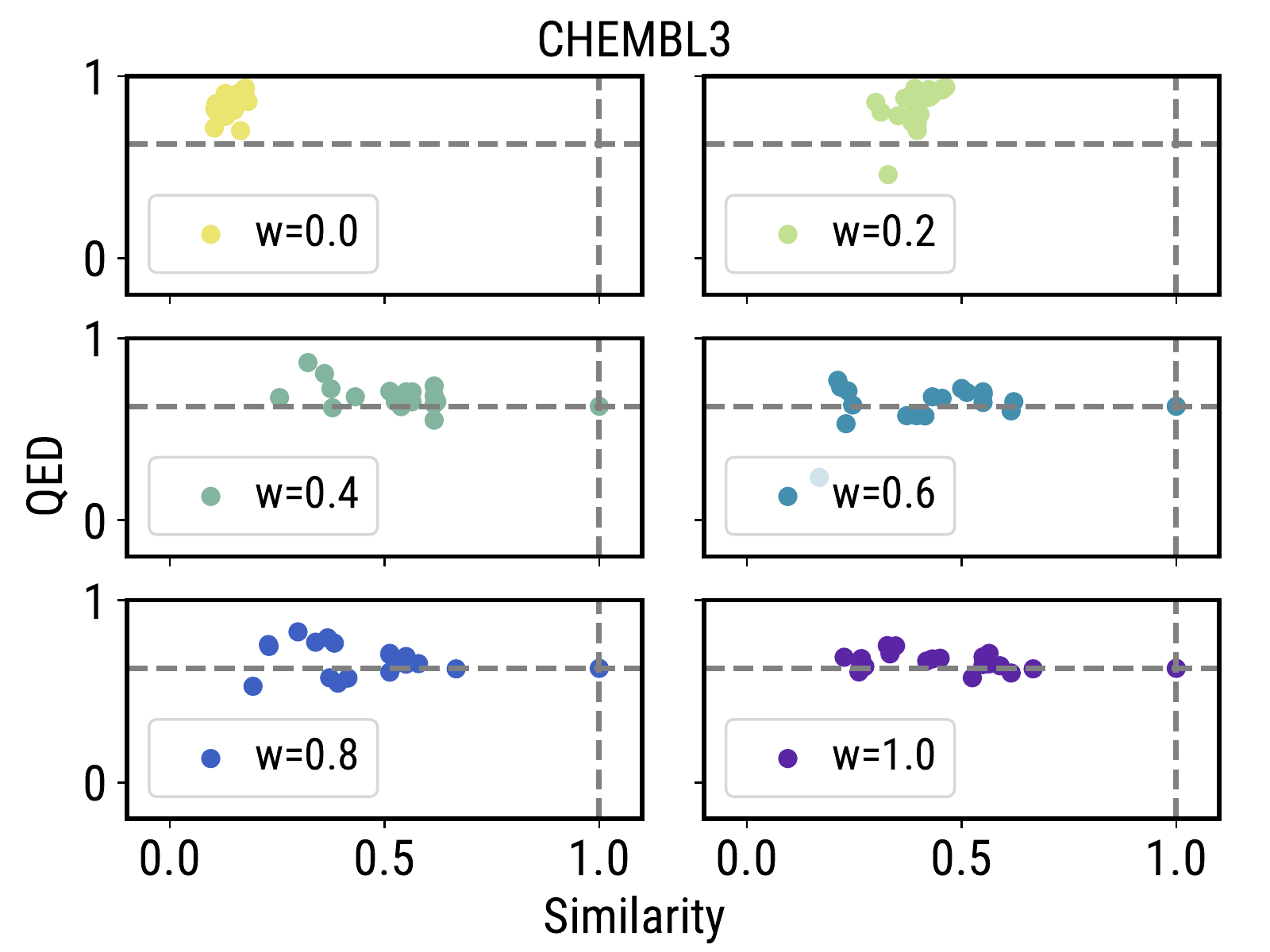} \\
  \includegraphics[width=0.5\textwidth]{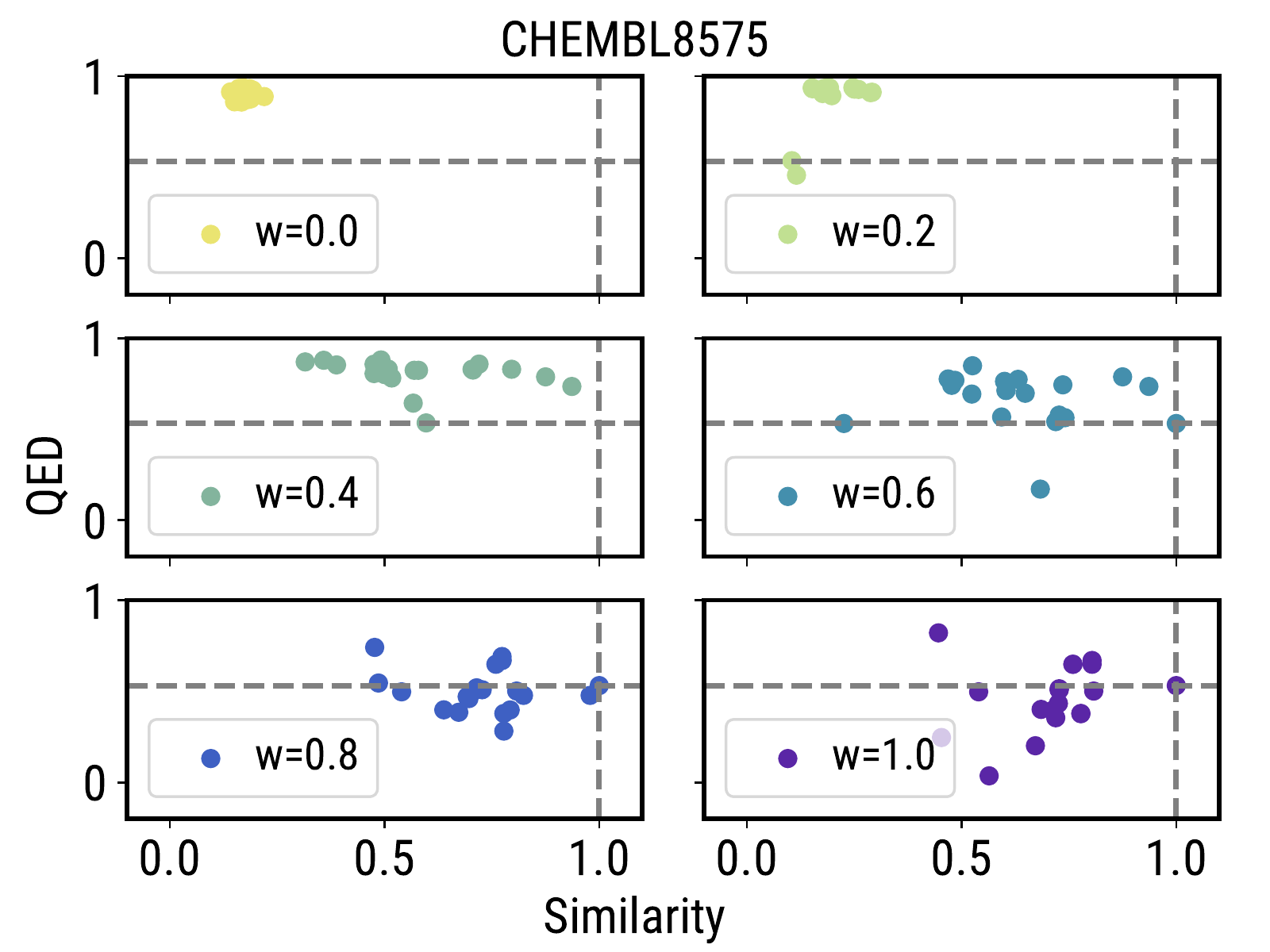} & 
  \includegraphics[width=0.5\textwidth]{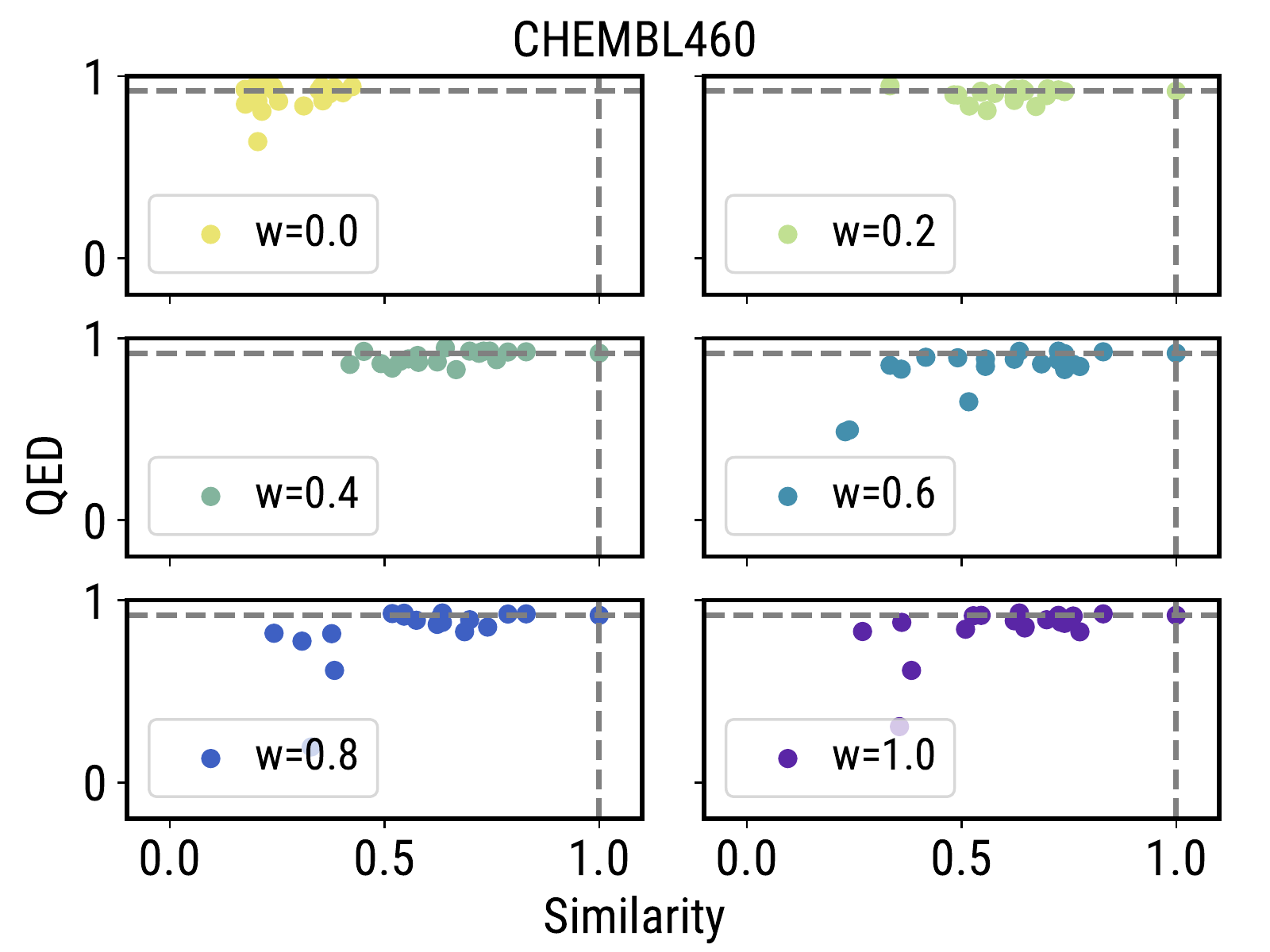} \\
  \includegraphics[width=0.5\textwidth]{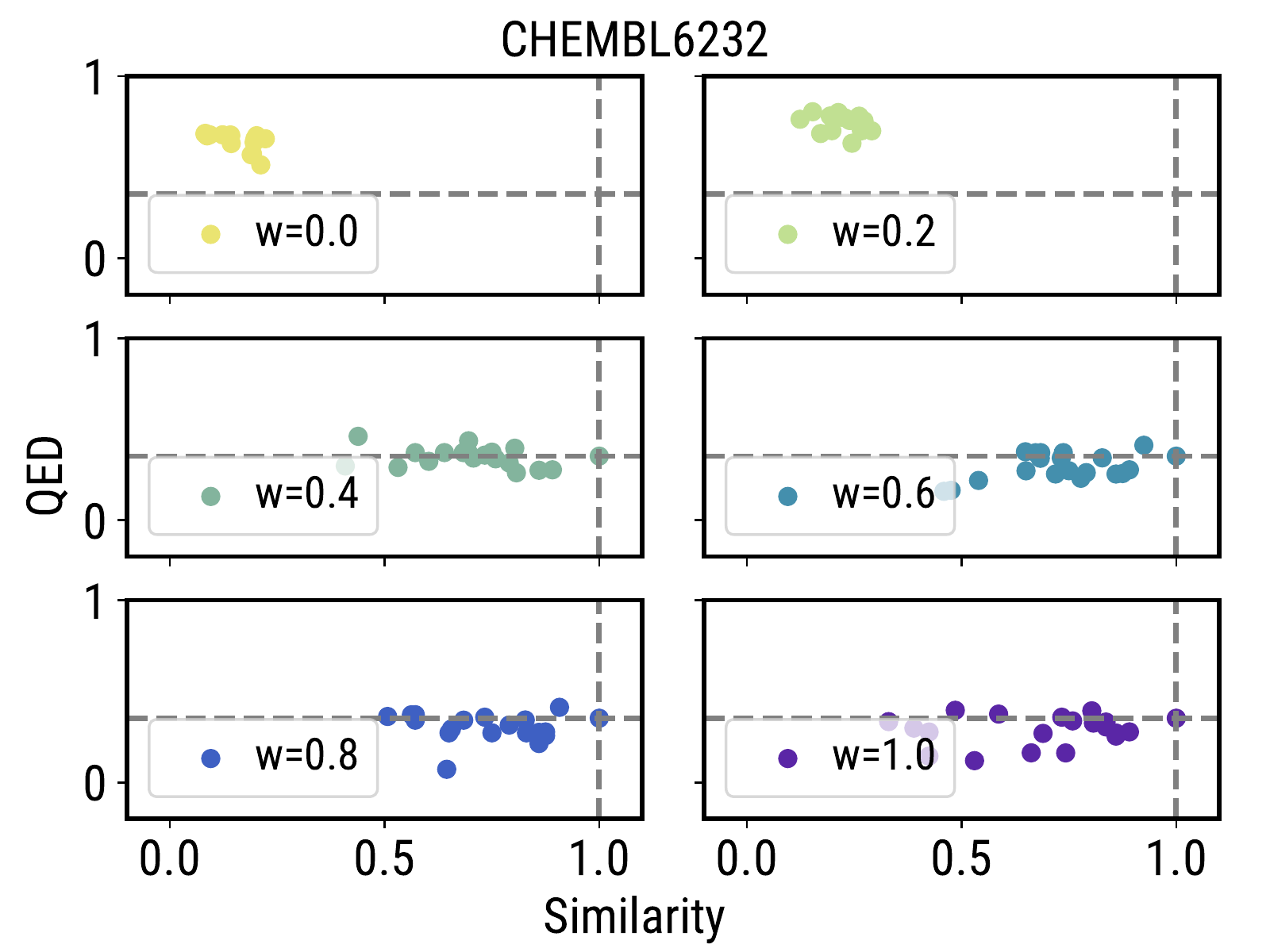} & 
  \includegraphics[width=0.5\textwidth]{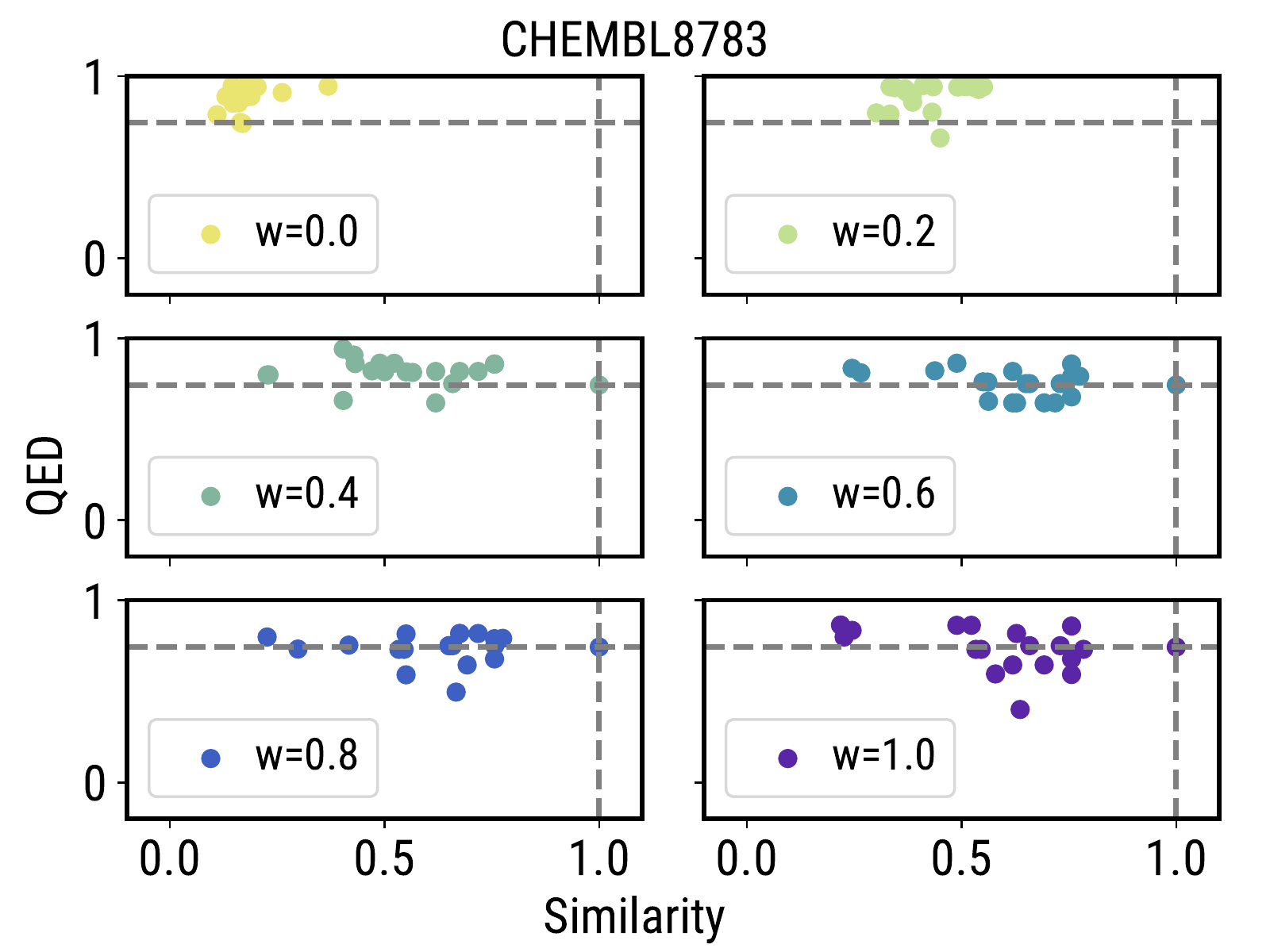} \\
\end{tabular}
\end{figure*}

\begin{figure*}
\begin{tabular}{ll}
  \includegraphics[width=0.5\textwidth]{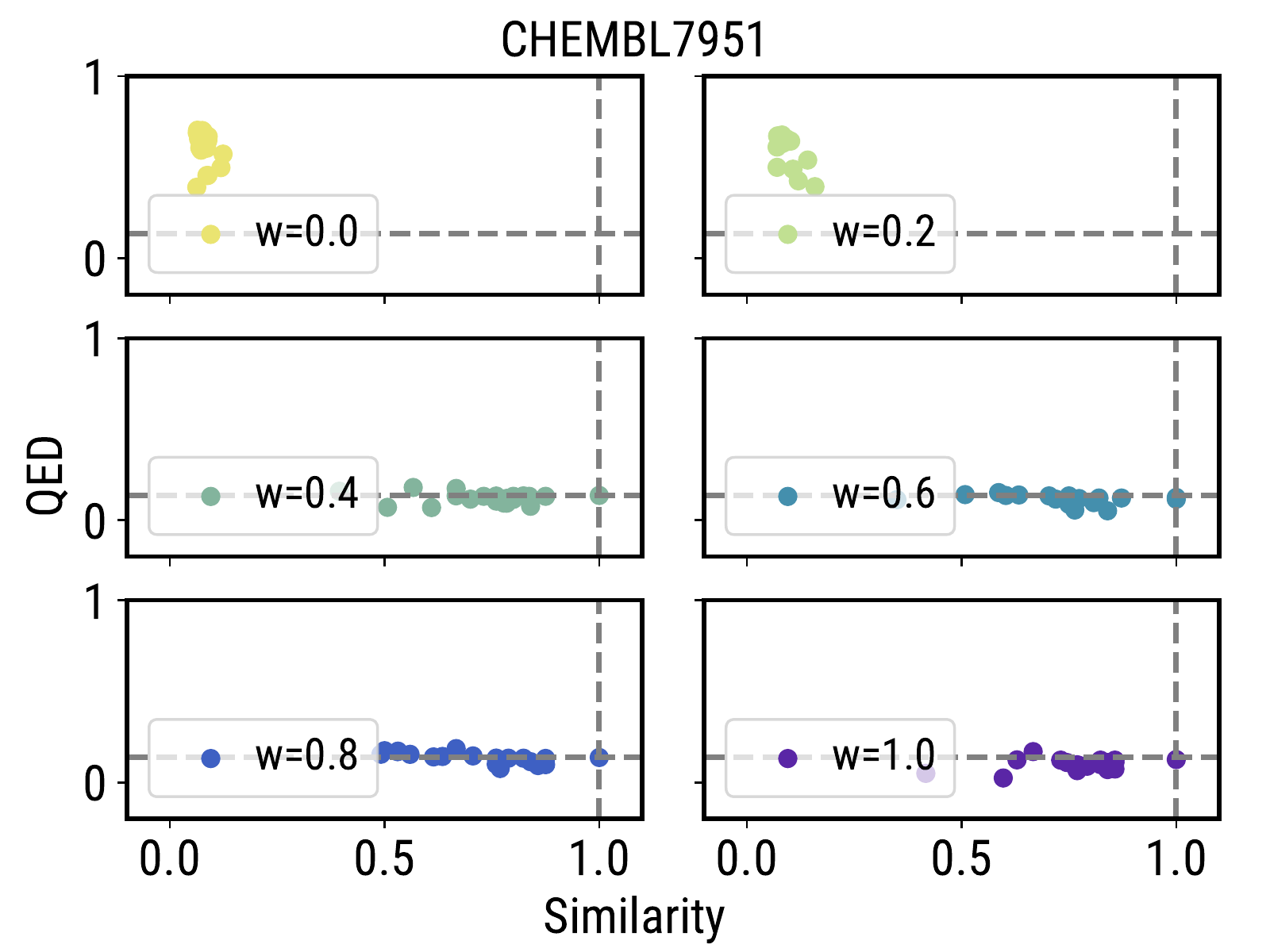} & 
  \includegraphics[width=0.5\textwidth]{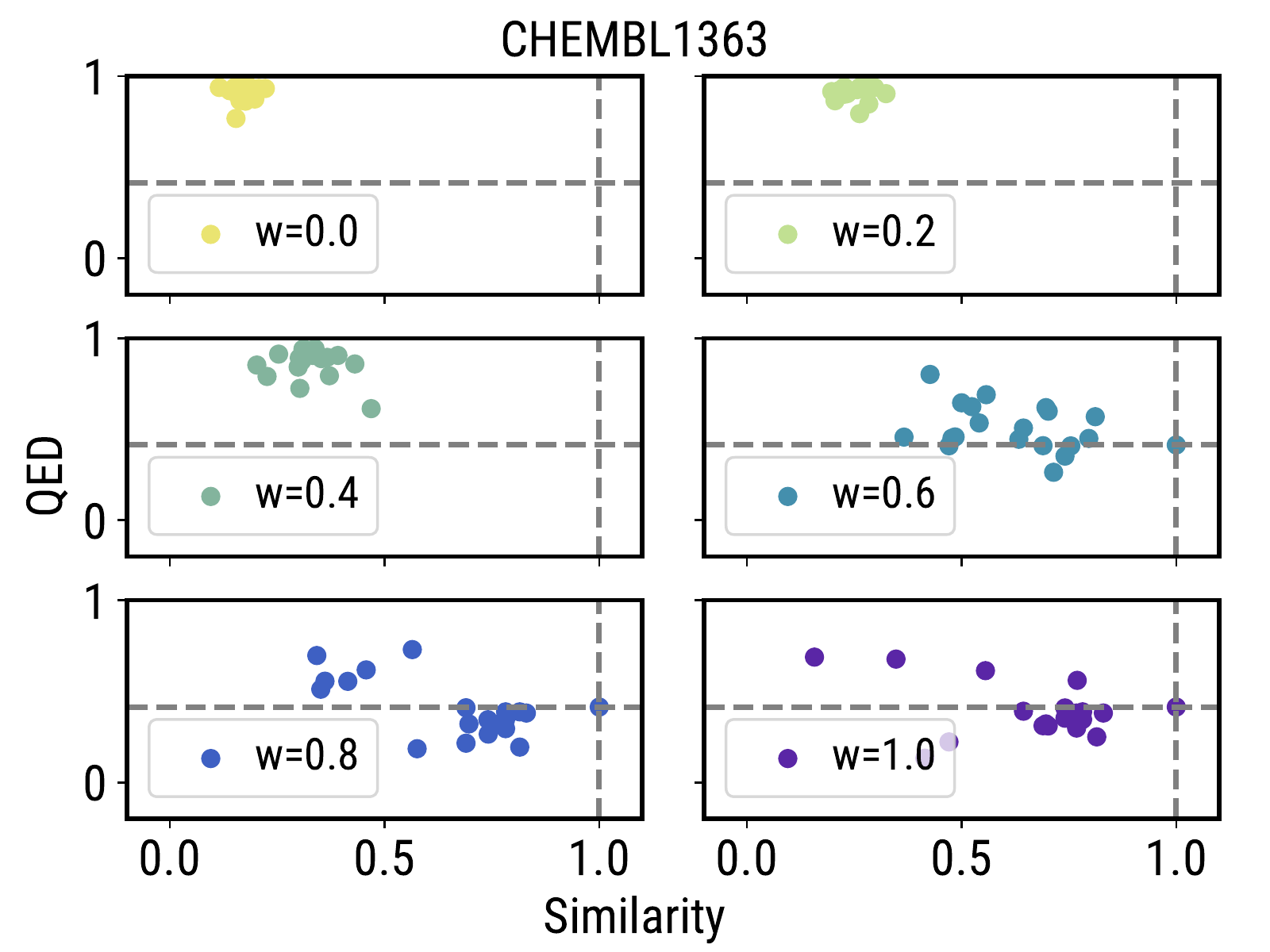} \\
  \includegraphics[width=0.5\textwidth]{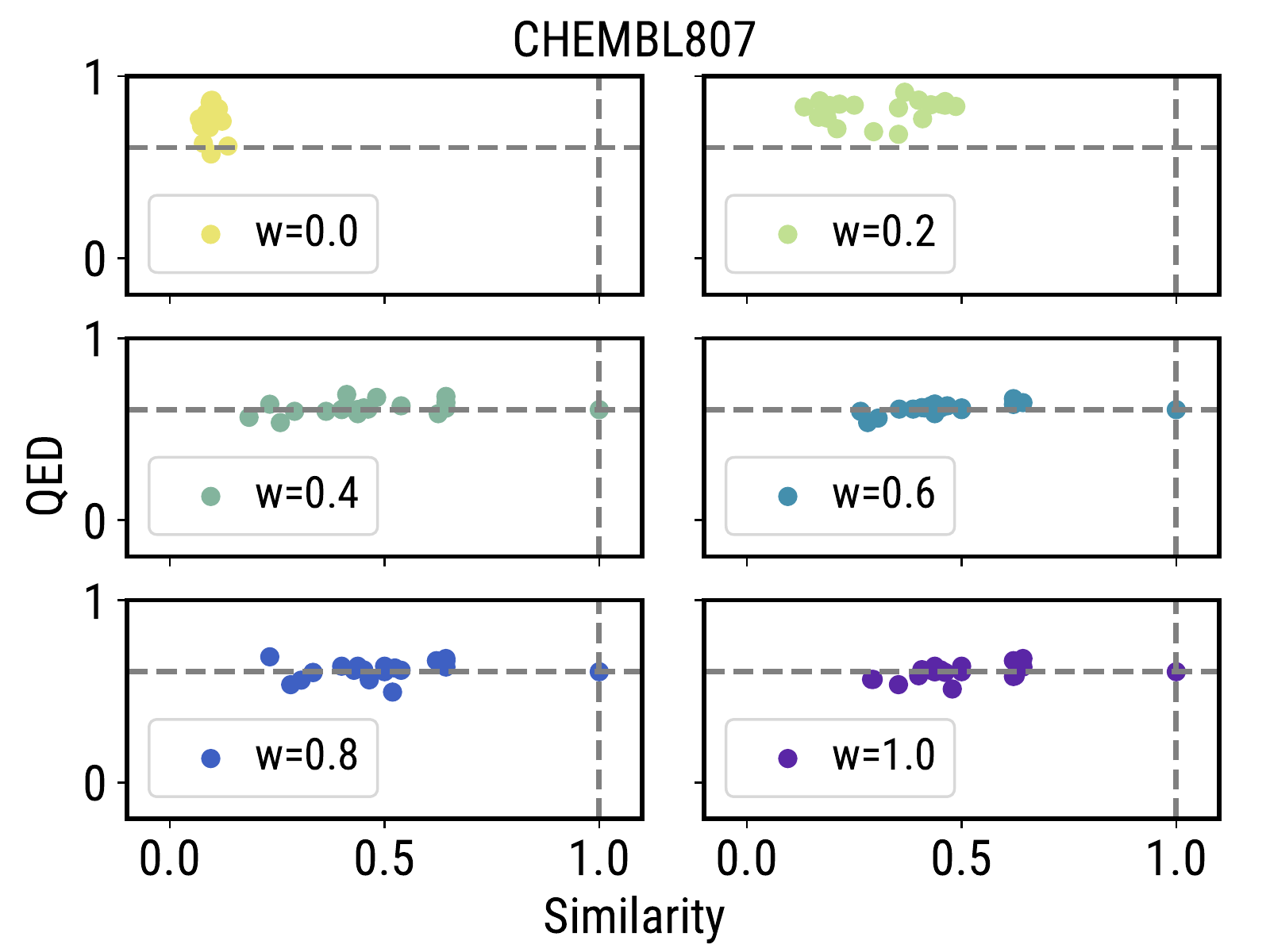} & 
  \includegraphics[width=0.5\textwidth]{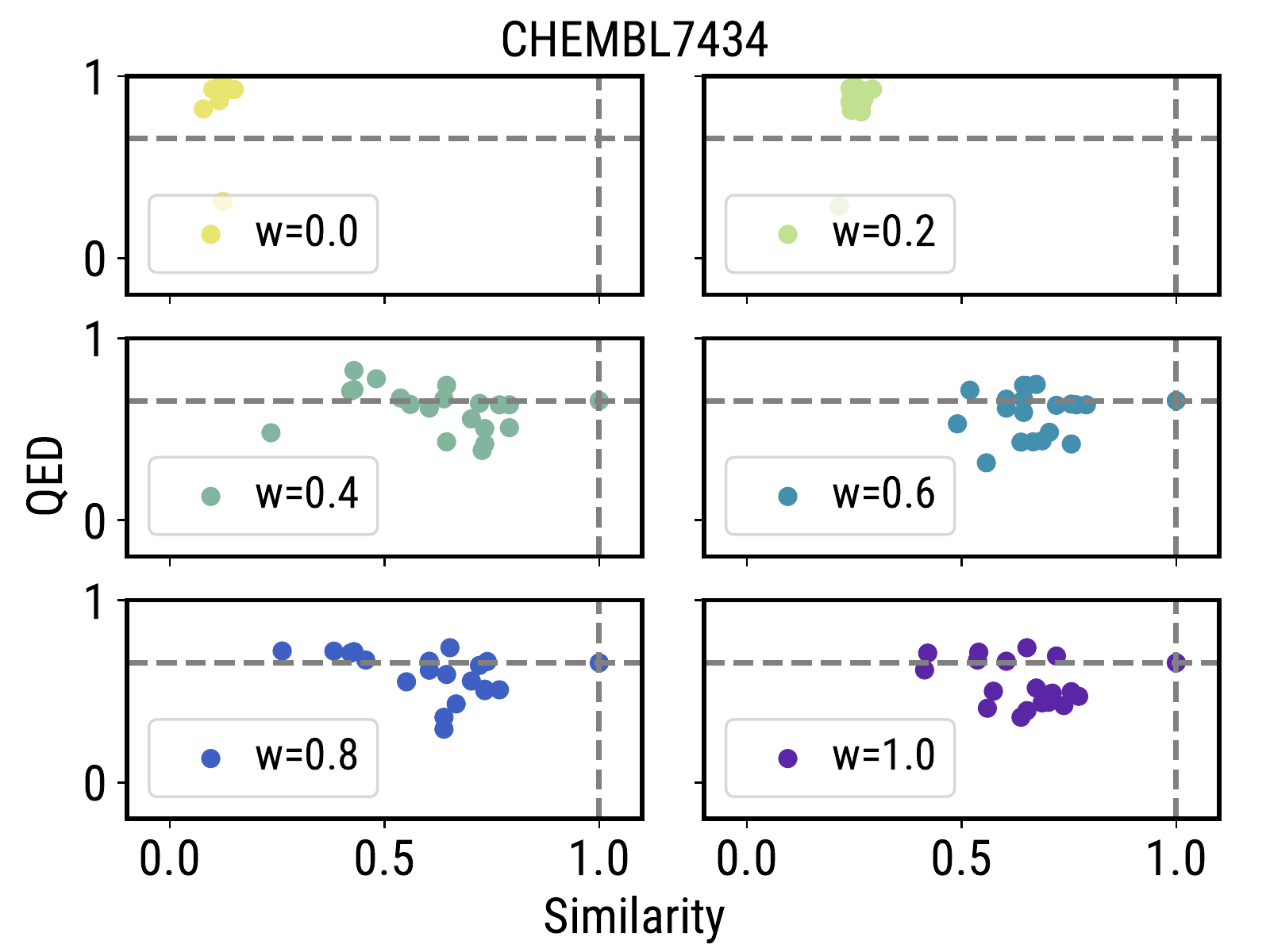} \\
  \includegraphics[width=0.5\textwidth]{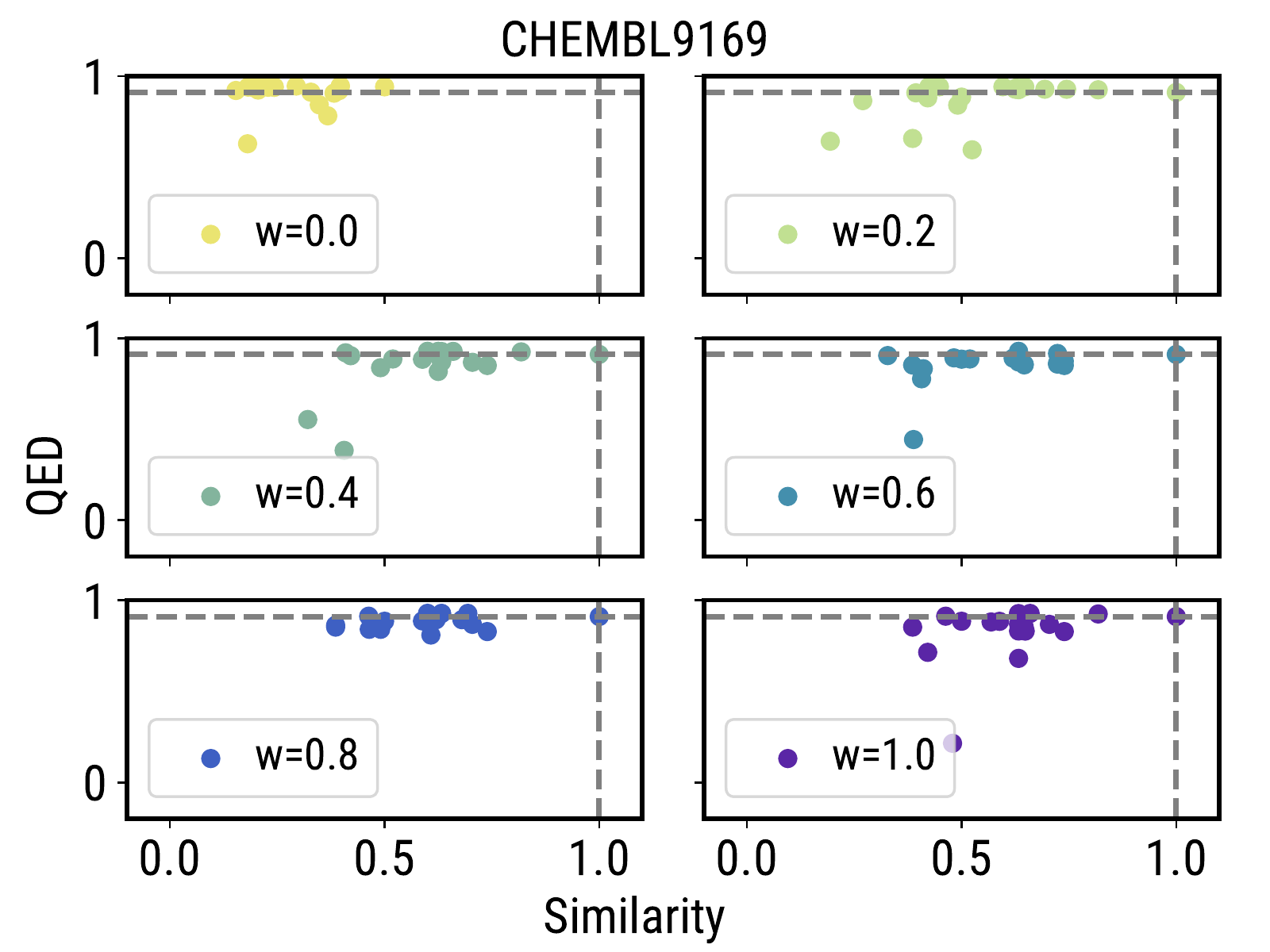} & 
  \includegraphics[width=0.5\textwidth]{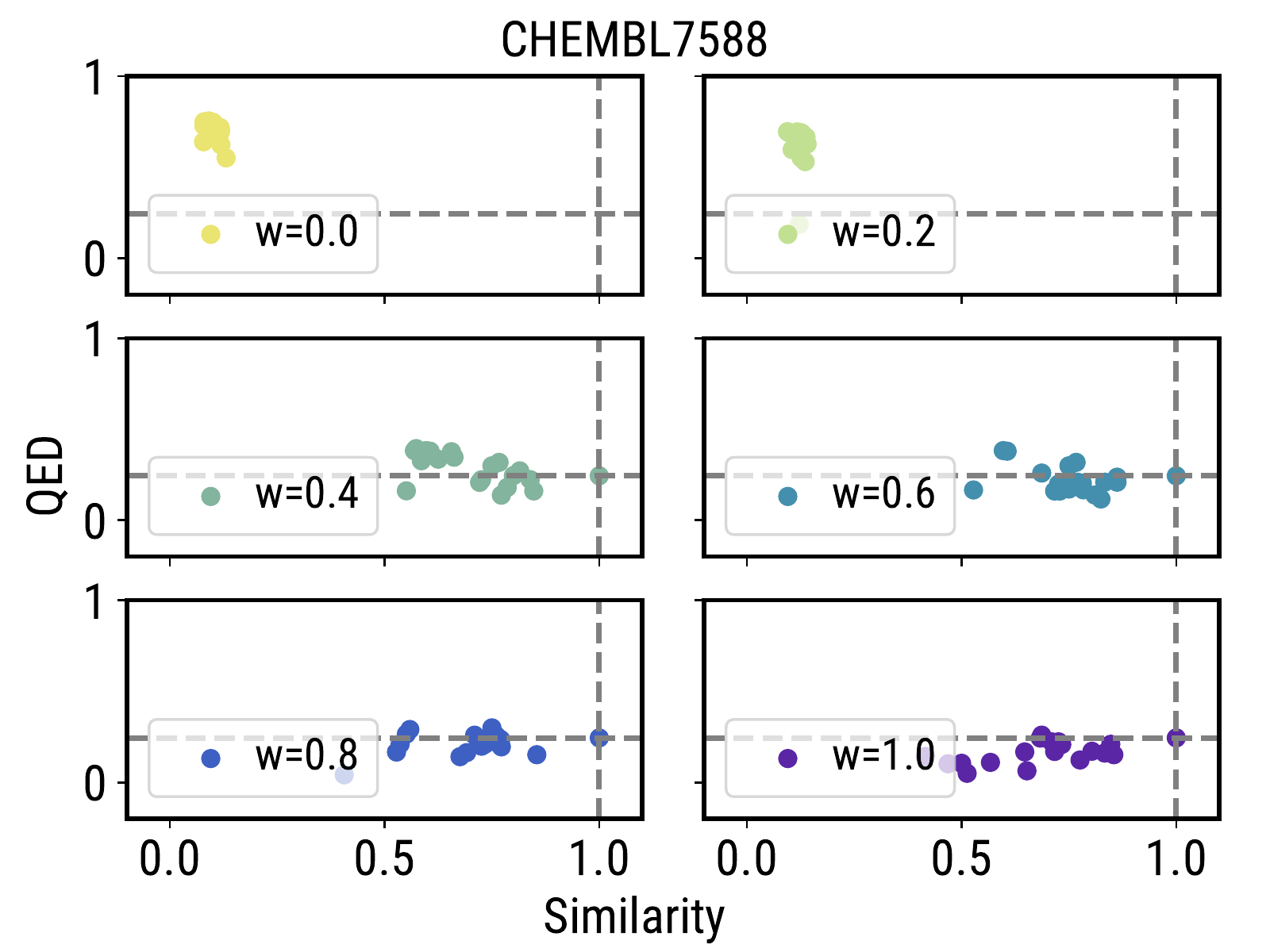} \\
\end{tabular}
\end{figure*}

\begin{figure*}
\begin{tabular}{ll}
  \includegraphics[width=0.5\textwidth]{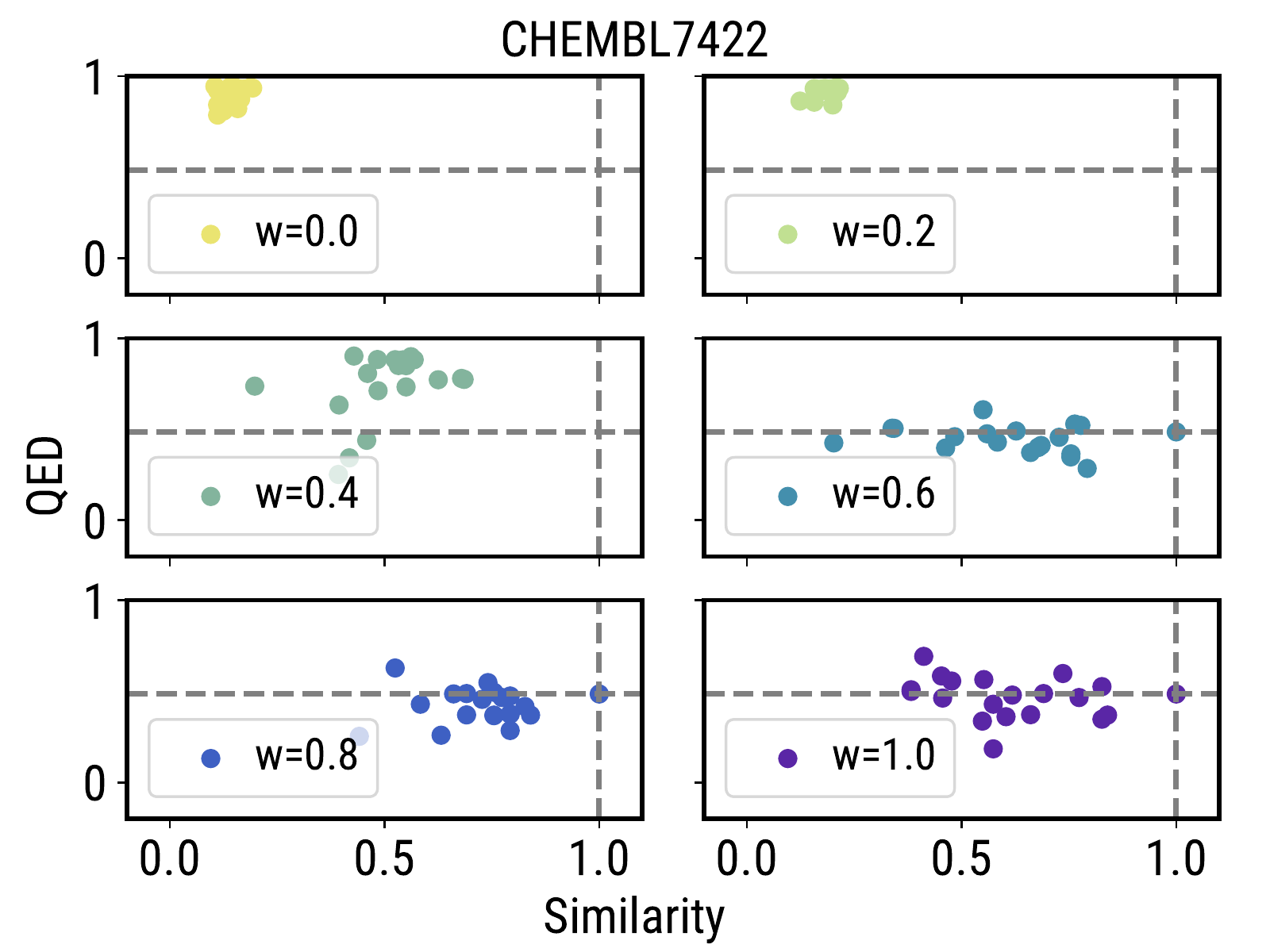} & 
  \includegraphics[width=0.5\textwidth]{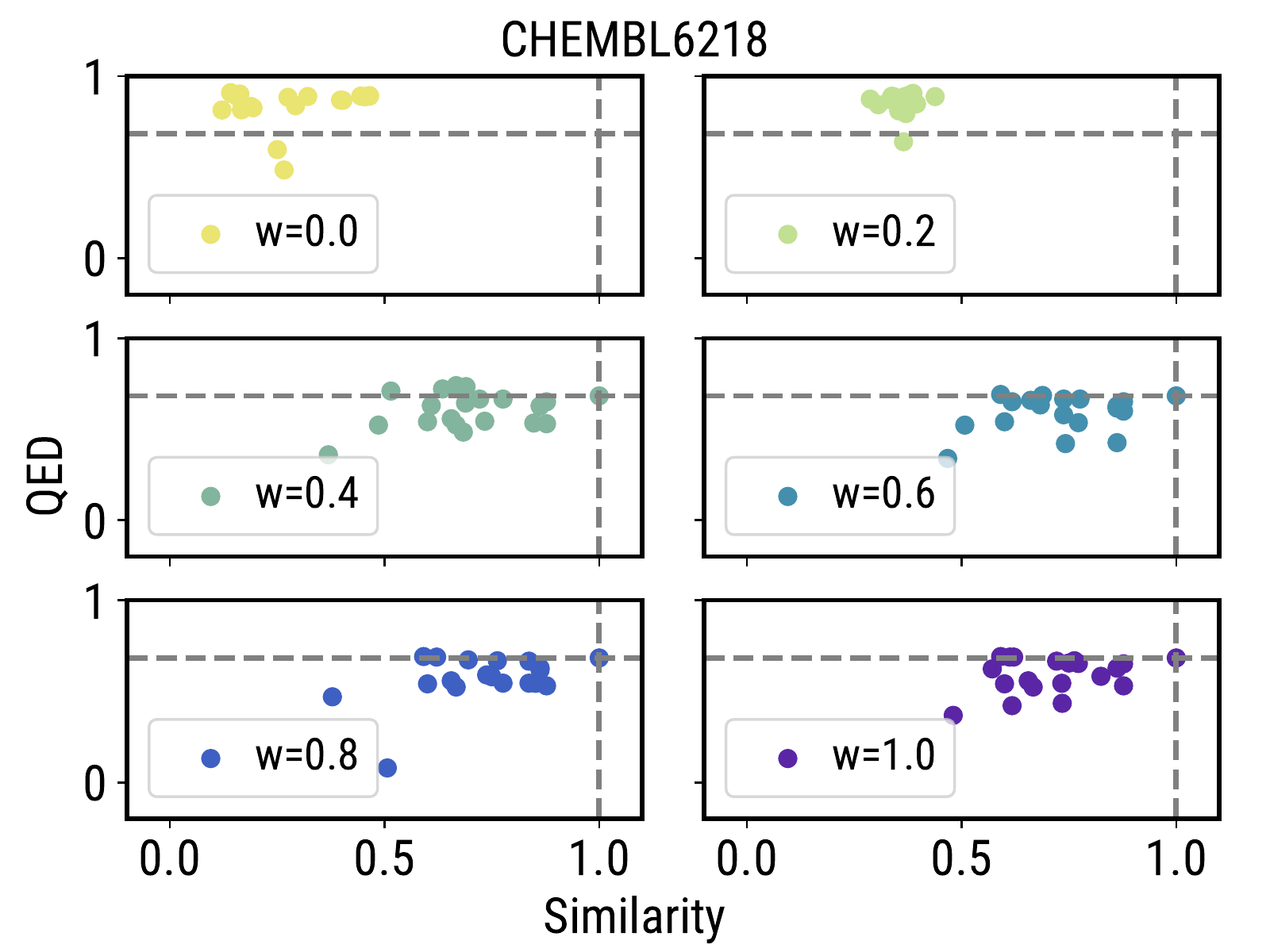} \\
  \includegraphics[width=0.5\textwidth]{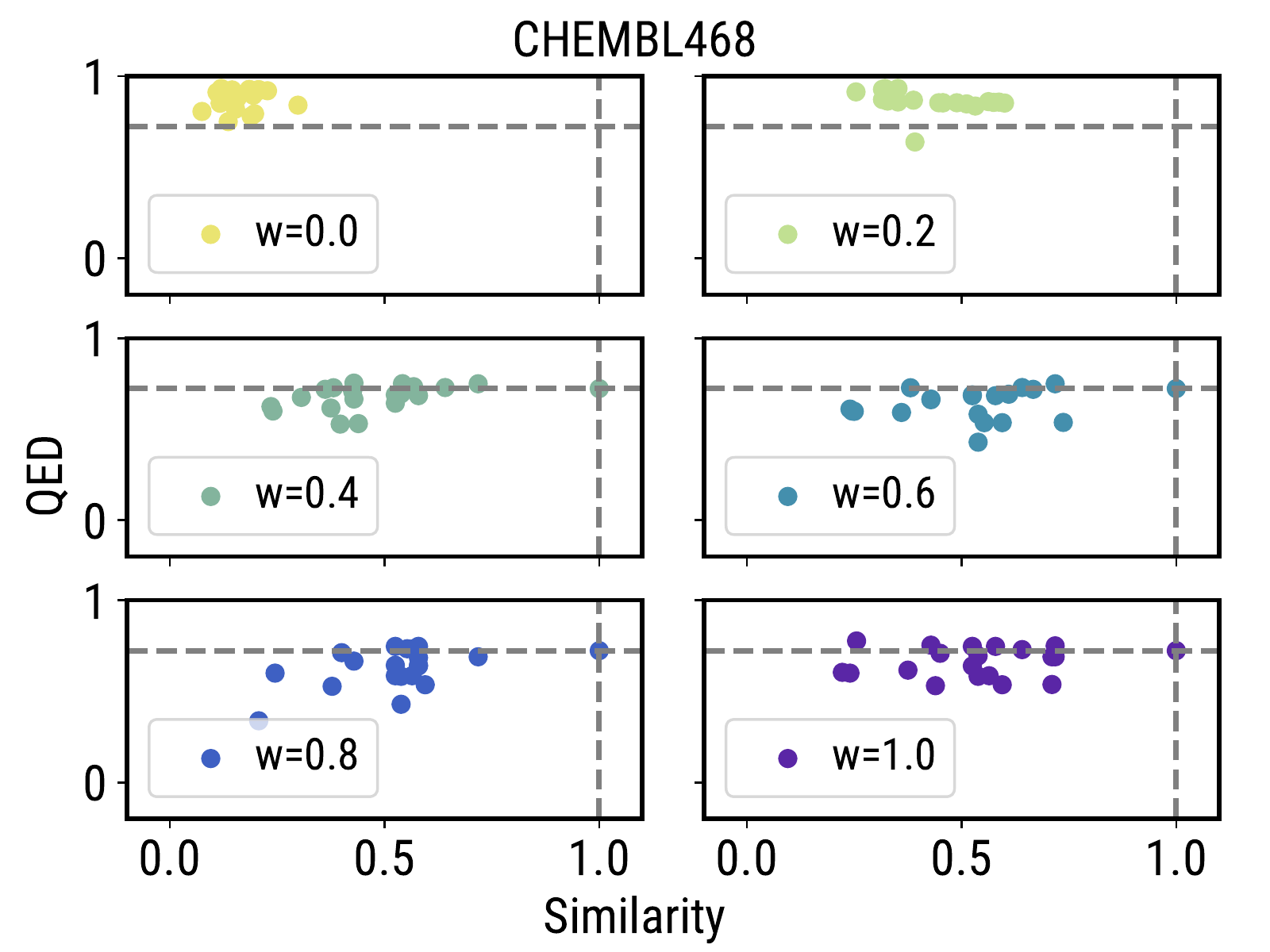} & 
  \includegraphics[width=0.5\textwidth]{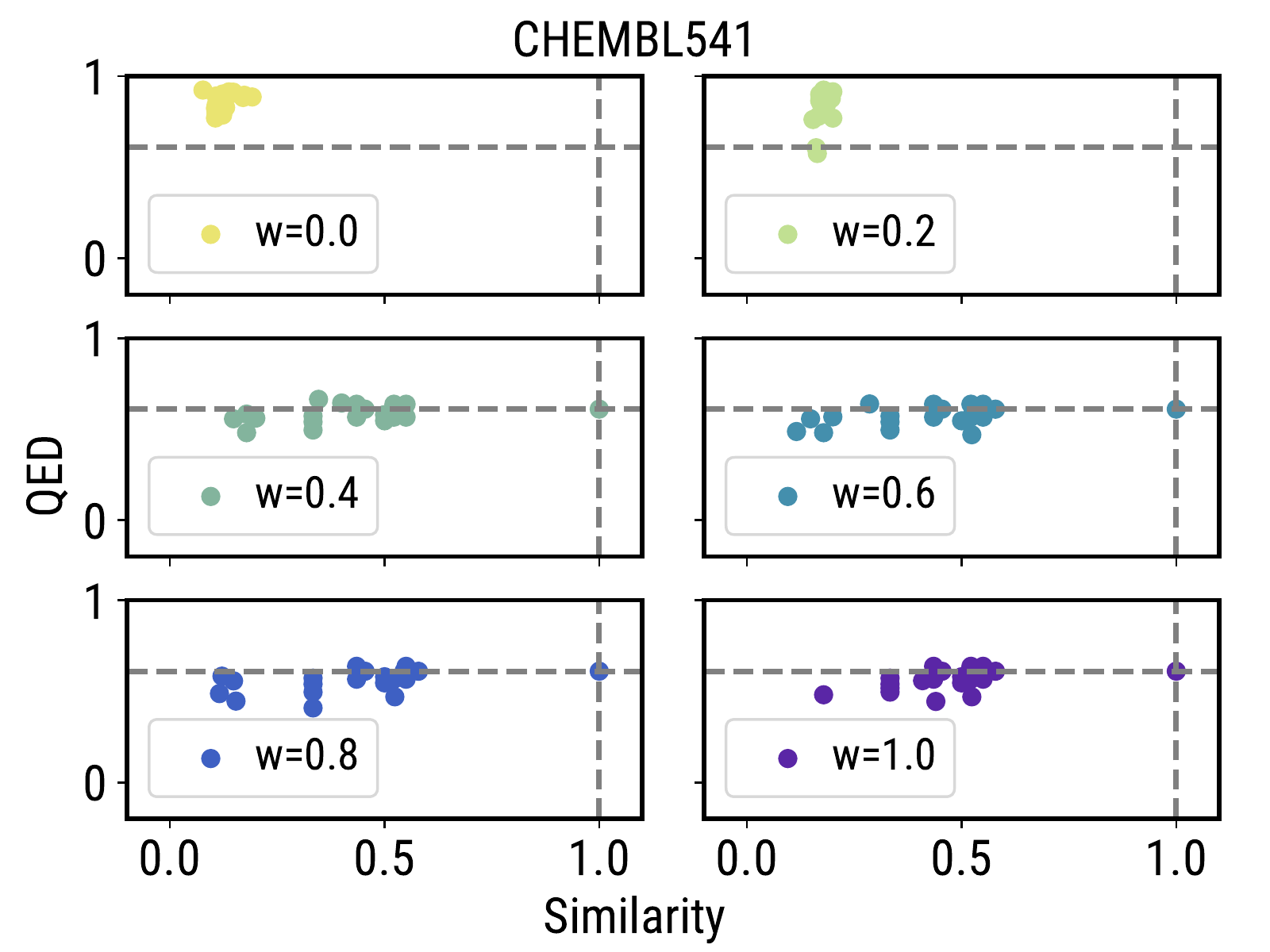} \\
  \includegraphics[width=0.5\textwidth]{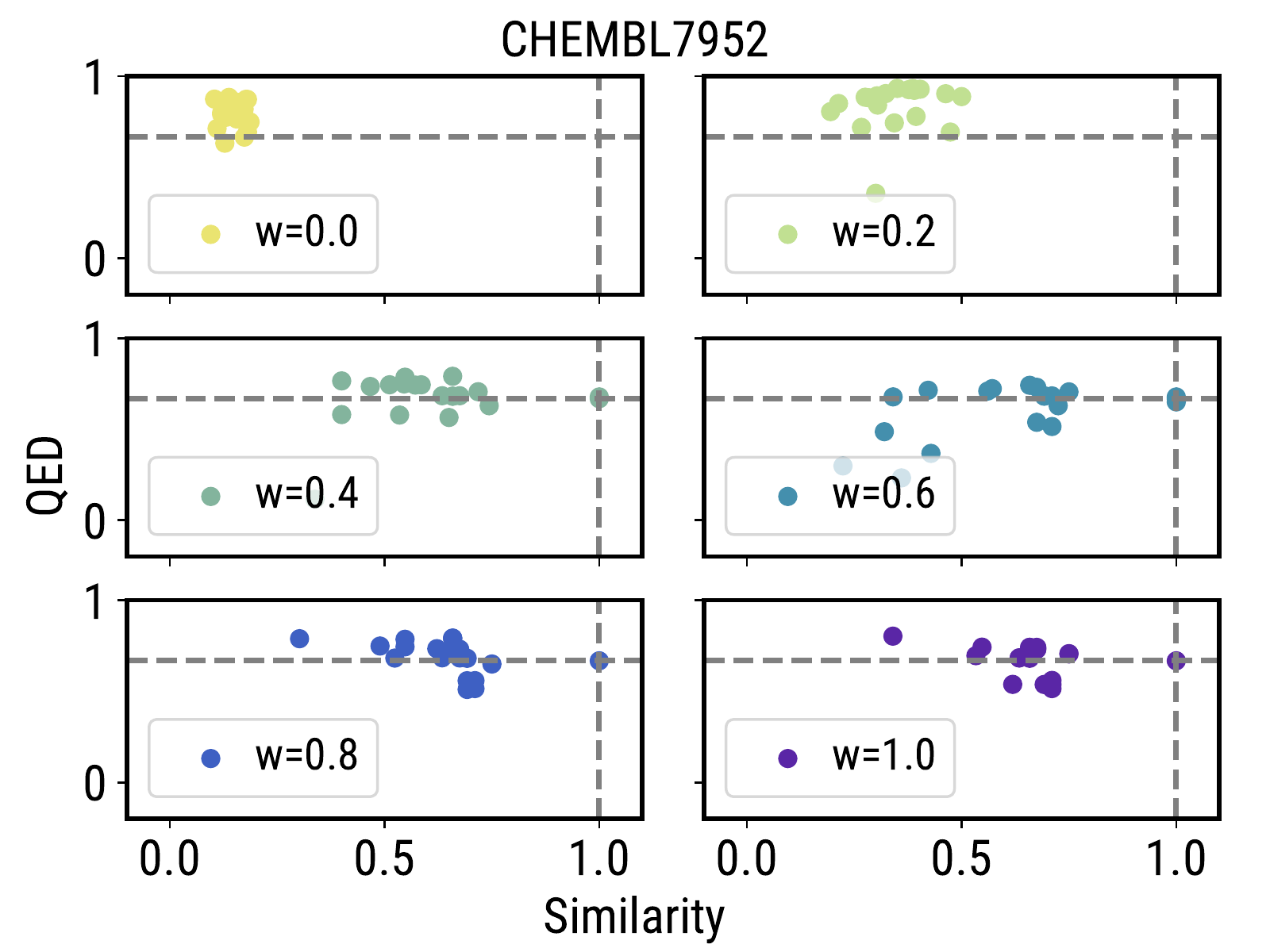} & 
  \includegraphics[width=0.5\textwidth]{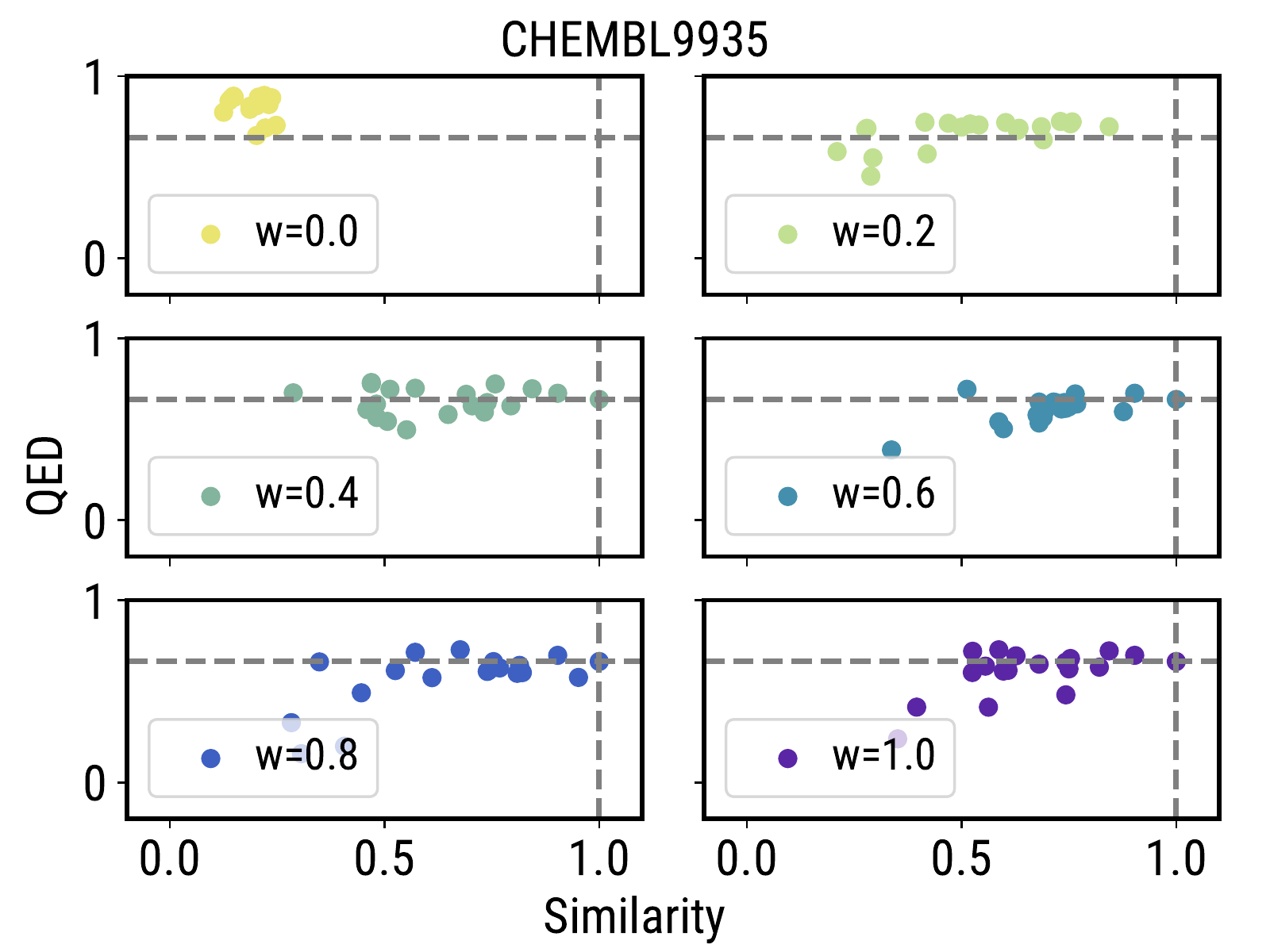} \\
\end{tabular}
\end{figure*}

\begin{figure}
\begin{tabular}{ll}
  \includegraphics[width=0.5\textwidth]{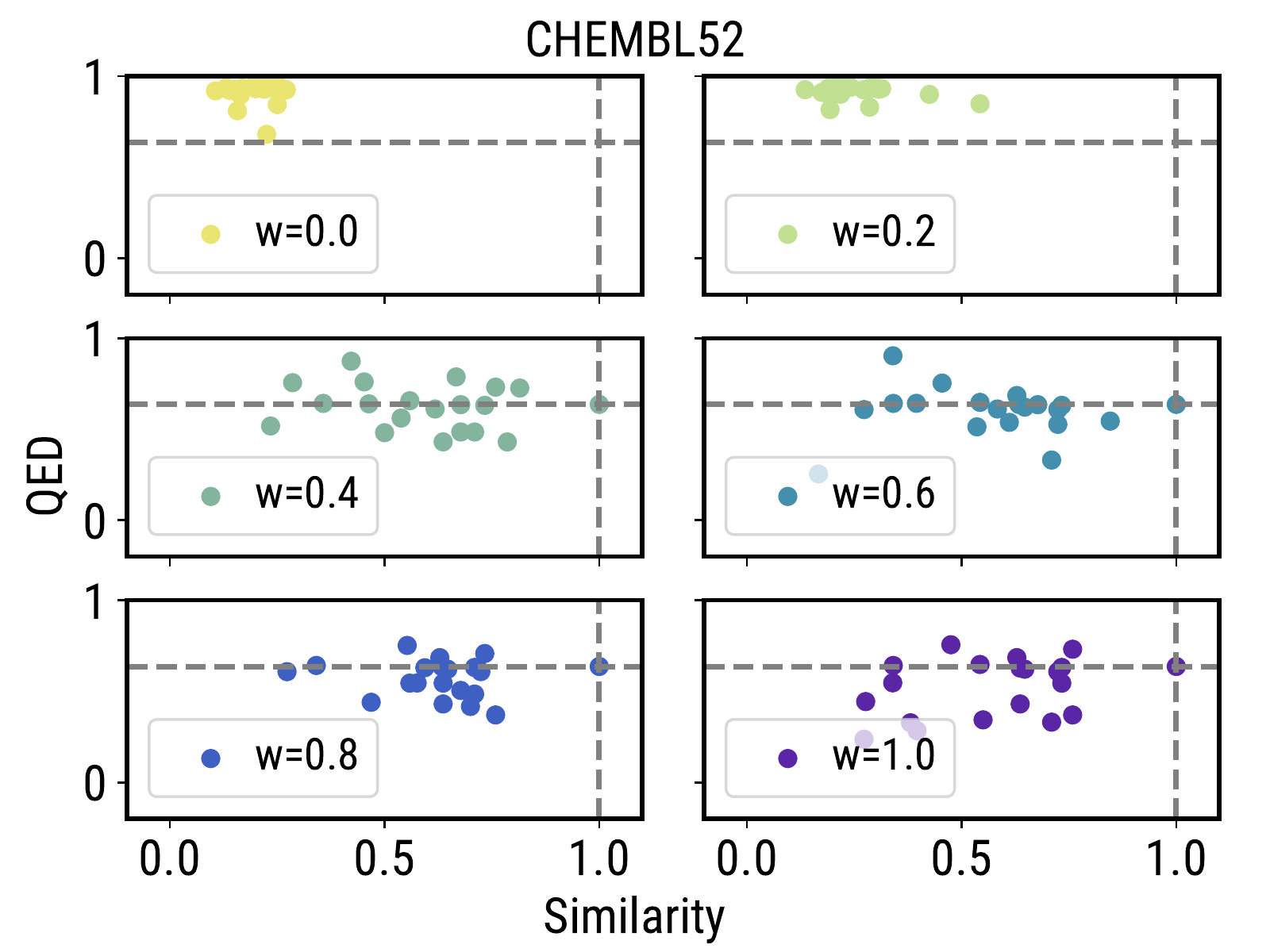} & 
  \includegraphics[width=0.5\textwidth]{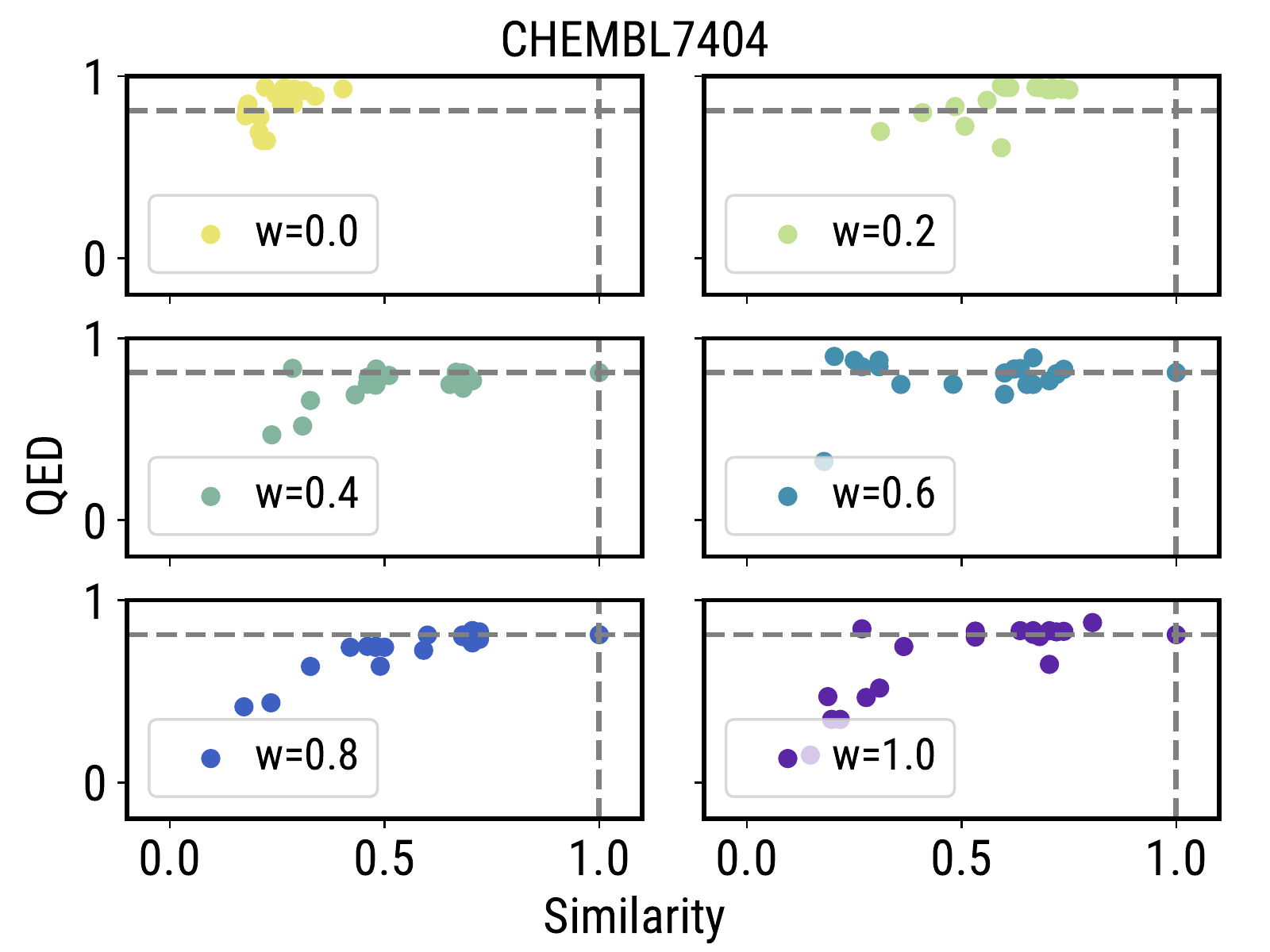} \\
  \includegraphics[width=0.5\textwidth]{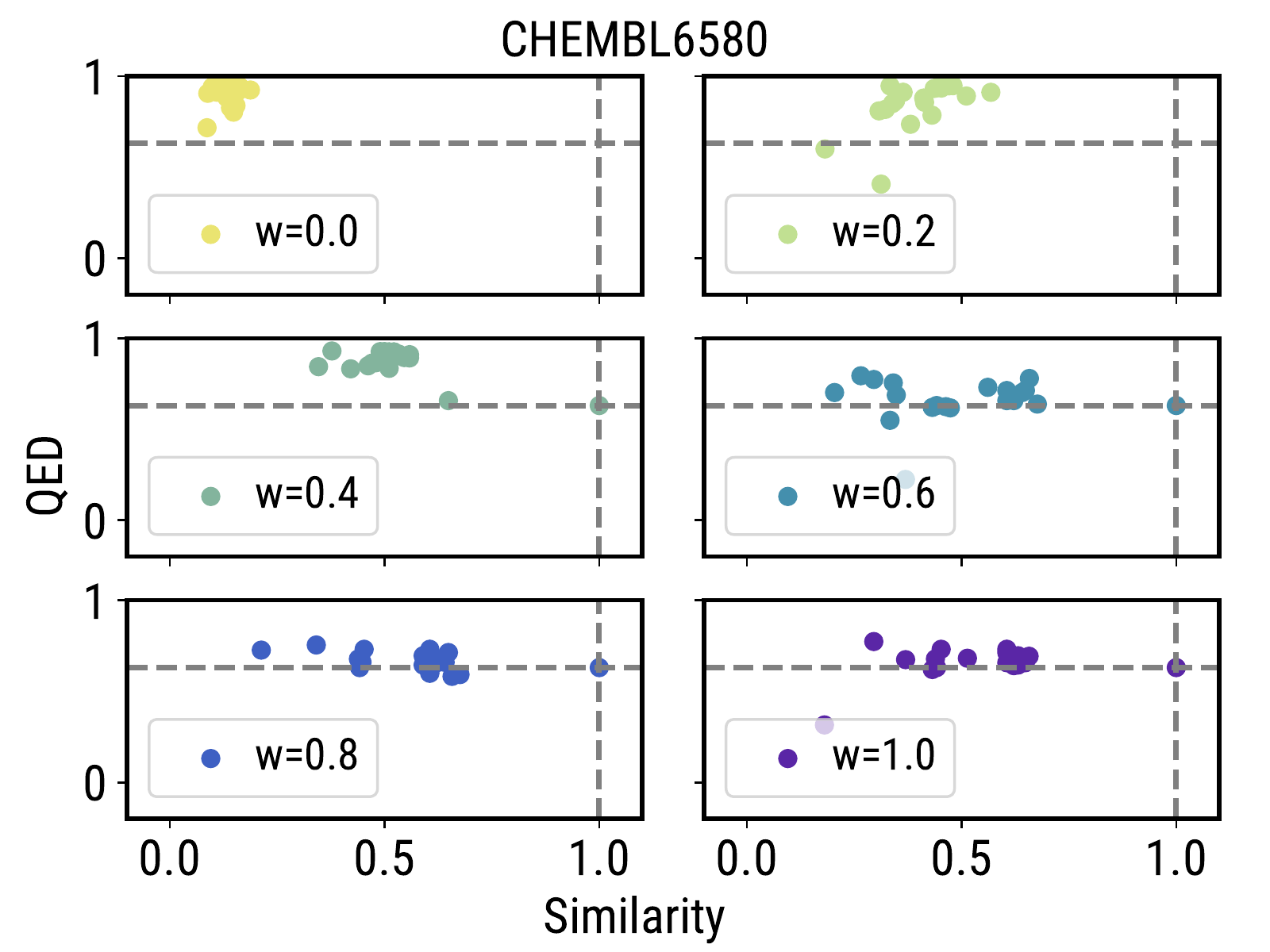} & 
  \includegraphics[width=0.5\textwidth]{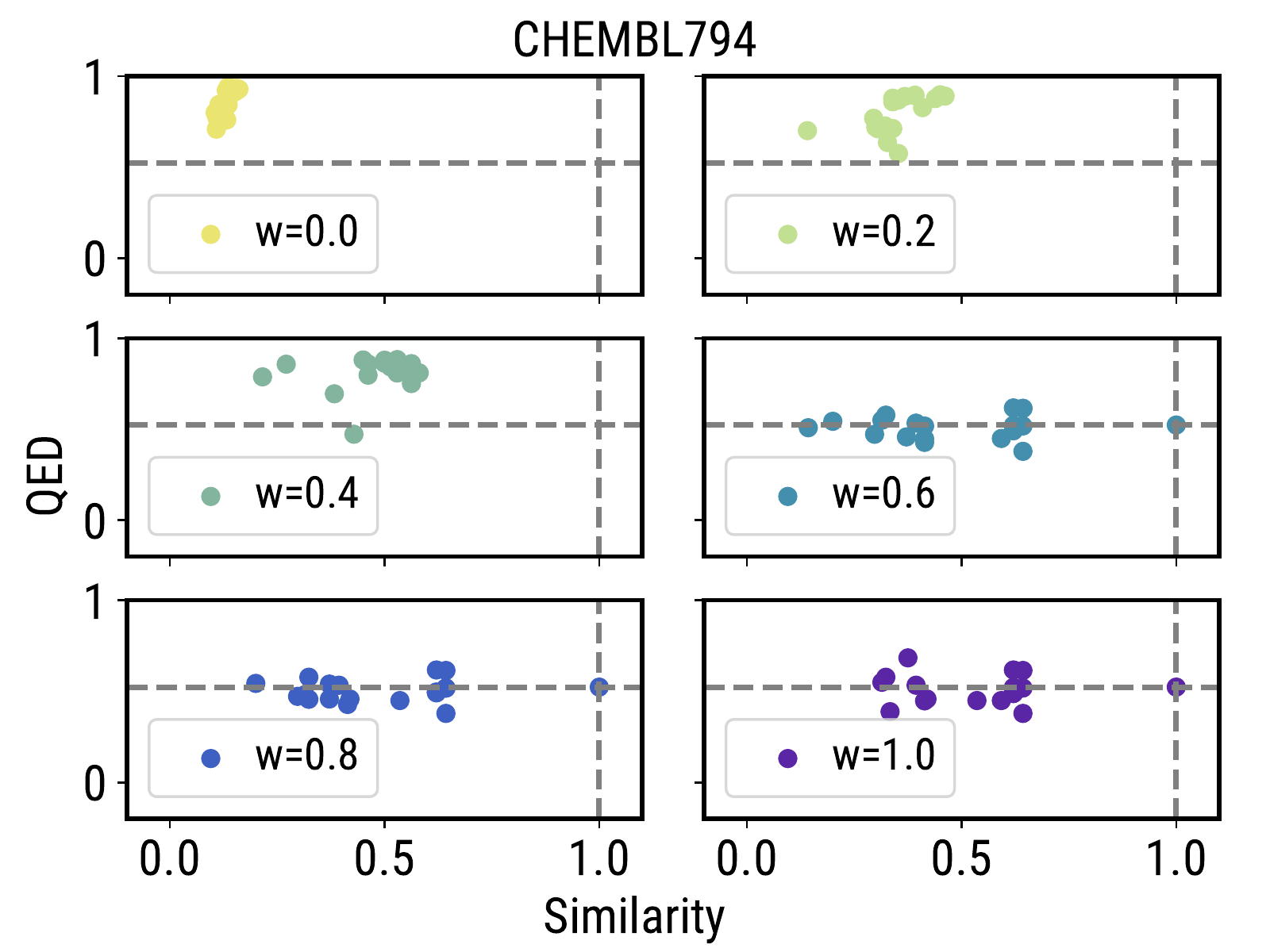} \\
  \includegraphics[width=0.5\textwidth]{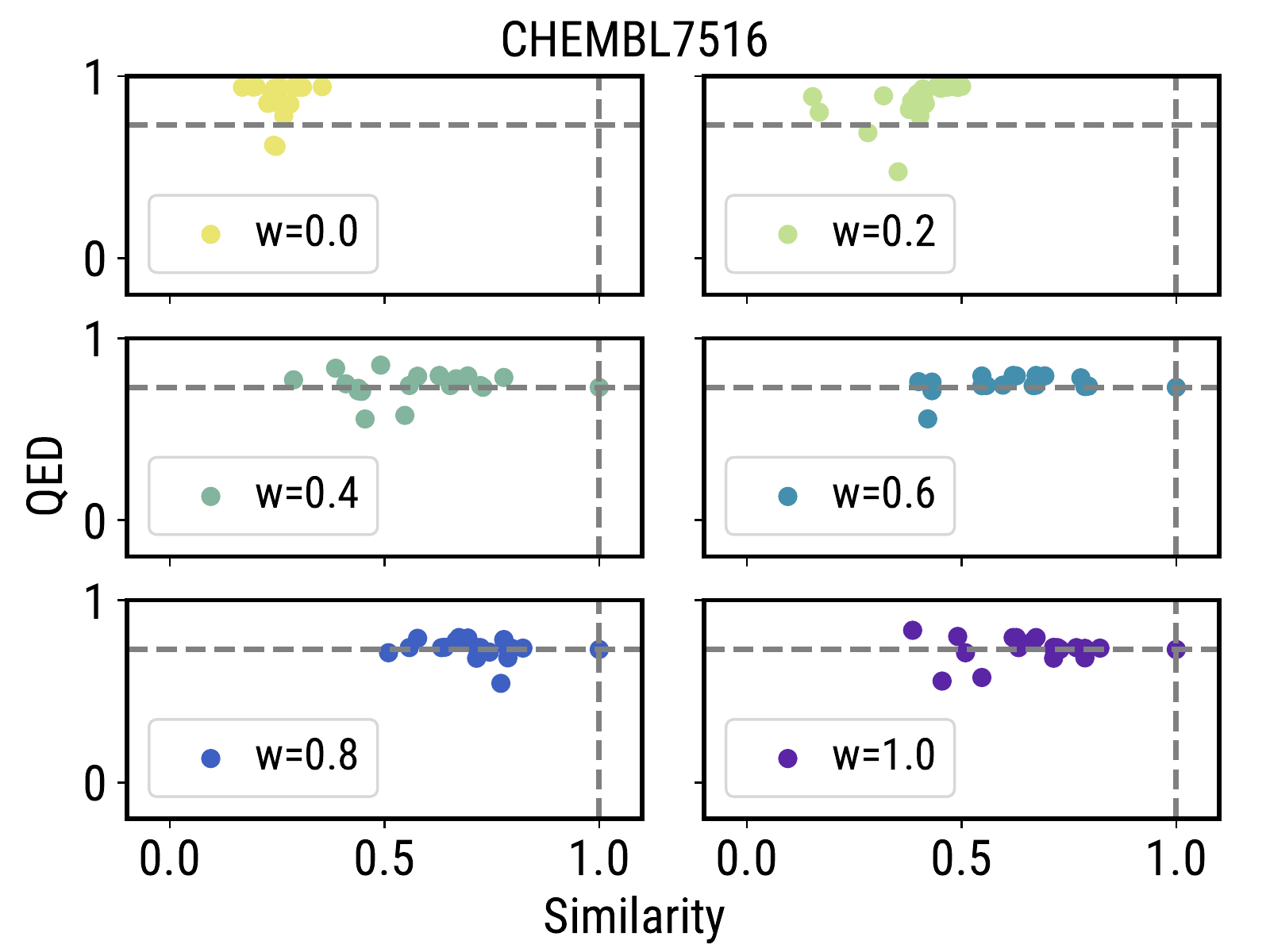} & 
  \includegraphics[width=0.5\textwidth]{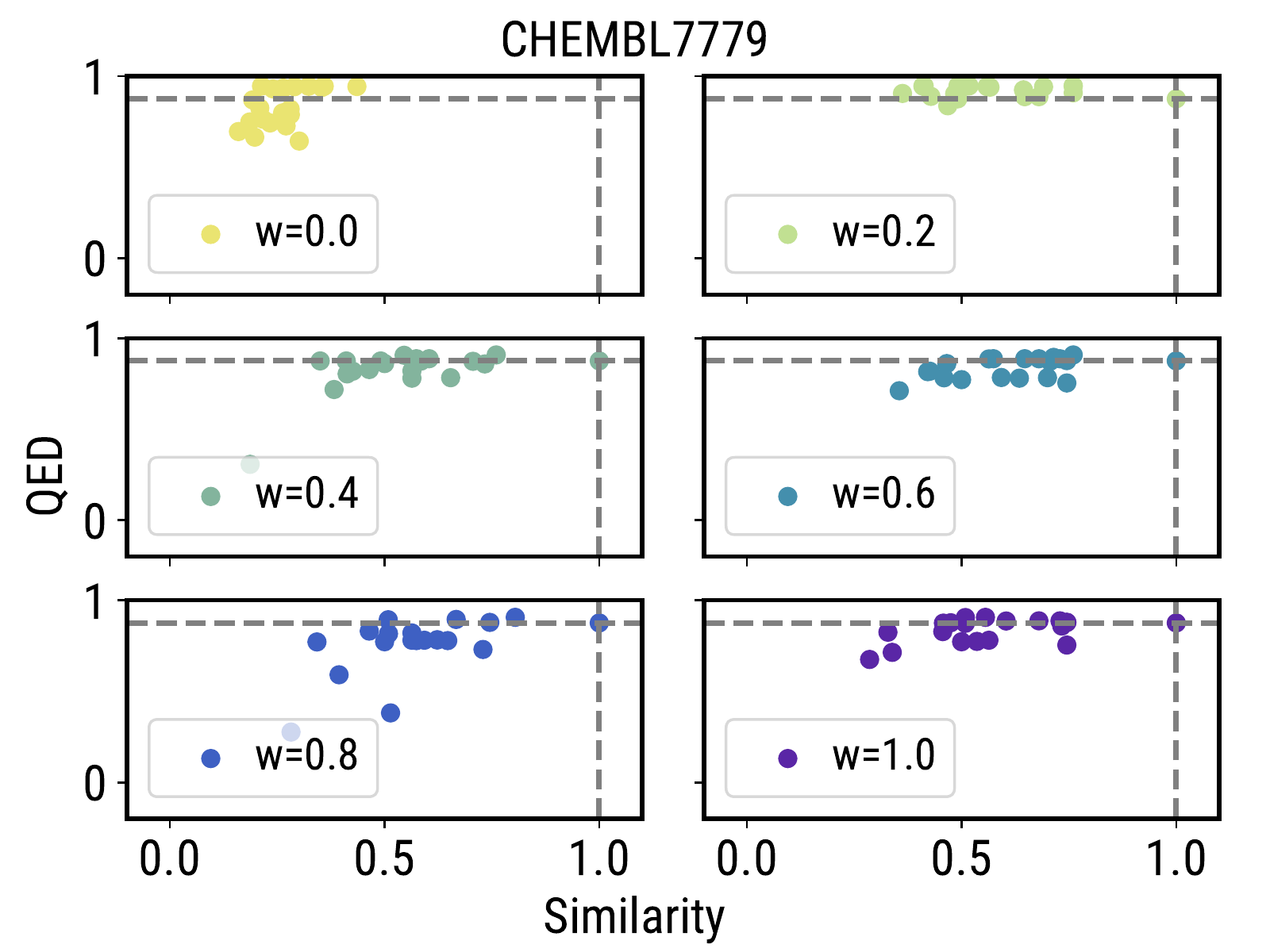} \\
\end{tabular}
  \caption{The QED and similarity of the molecules generated 
  under different weights with different staring molecules. 
  The gray dash line shows the QED and 
  similarity score of the starting molecule. }
  \label{fg:mult_obj_si_3}
\end{figure}
\begin{figure}

  \includegraphics[width=0.8\textwidth]{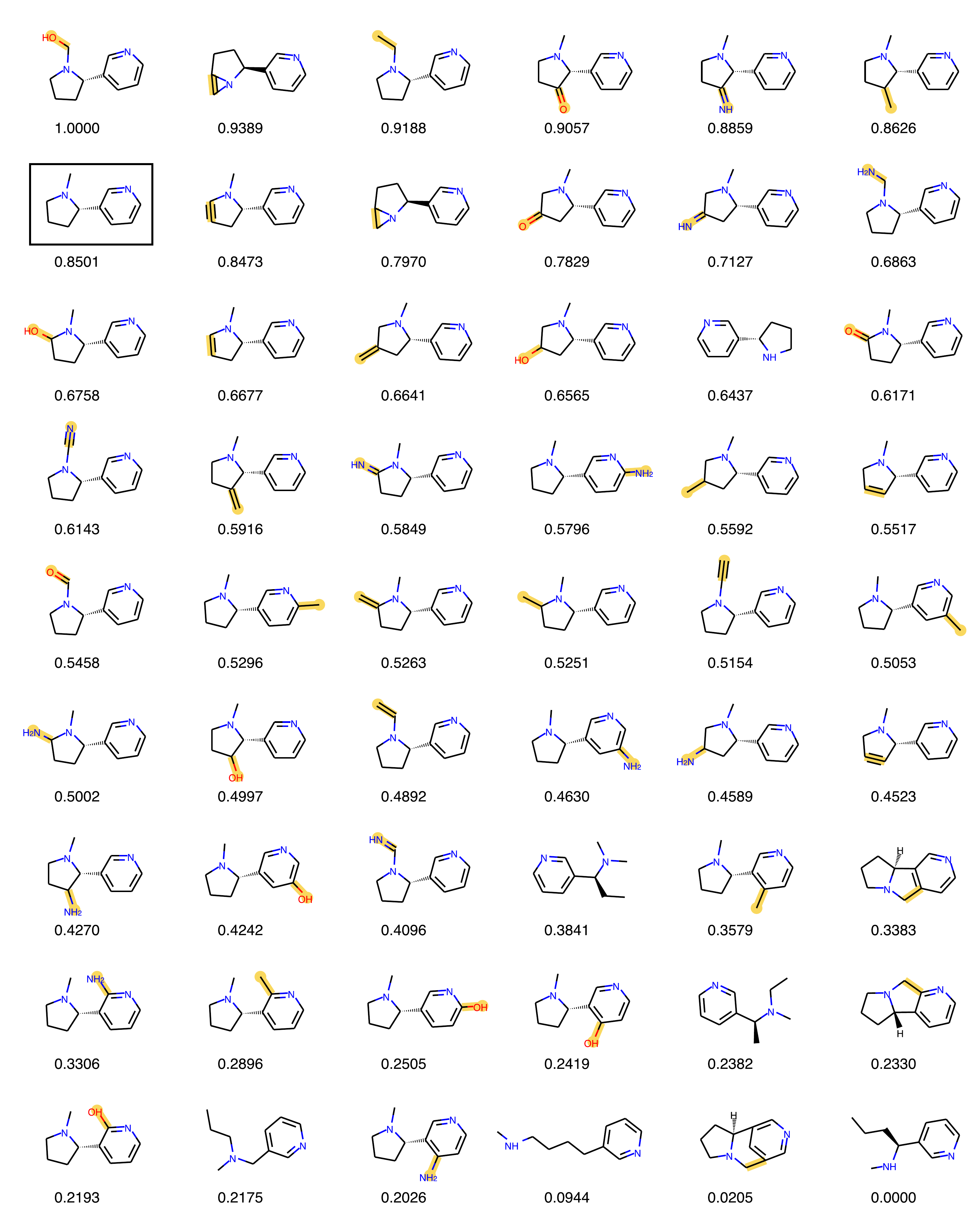} 

  \caption{The normalized $Q$-values of the actions can be
  taken in the first step. The original molecule was boxed. Bond addition actions are highlighted while bond removals are presented as is. The $Q$-values are rescaled to $[0, 1]$}
  \label{fg:qval_mat}
\end{figure}


\begin{figure}

  \includegraphics[width=\textwidth]{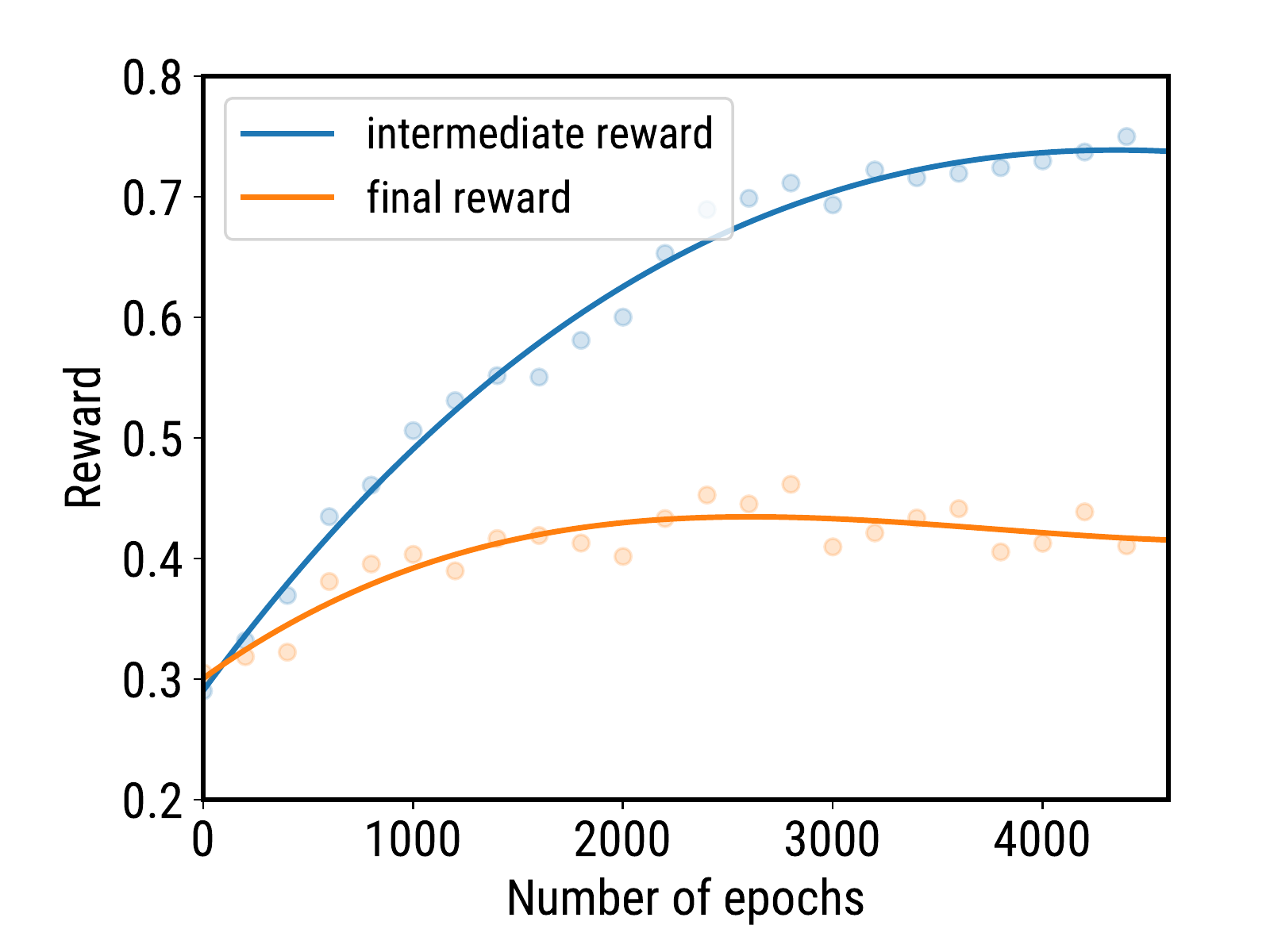} 

  \caption{Comparison between the learning curve of the agent while intermediate
  reward is given and that when only final reward is given. Here reward is defined
  as the QED of the final molecule generated. Bootstrap is turned off in this experiment.}
  \label{fg:reward_desgin}
\end{figure}

\clearpage

\begin{figure}

  \includegraphics[width=\textwidth]{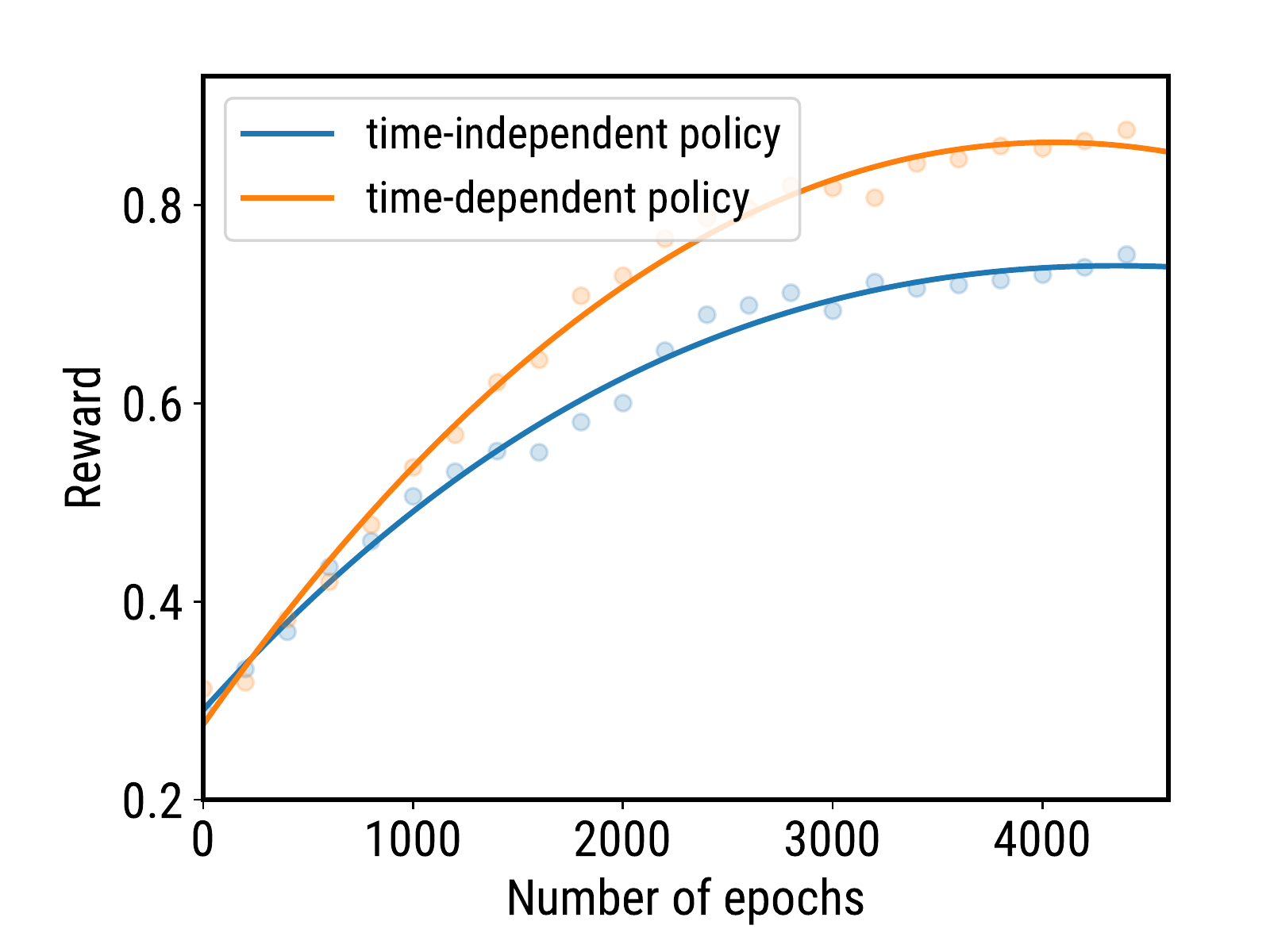} 

  \caption{Comparison of a time-dependent policy and
  a time-independent policy. Here reward is defined
  as the QED of the final molecule generated. Bootstrap is turned off in this experiment.}
  \label{fg:time_dep}
\end{figure}

\clearpage

\begin{figure}
    \begin{tabular}{cc}
    (a) & (b) \\
  \includegraphics[width=0.5\textwidth]{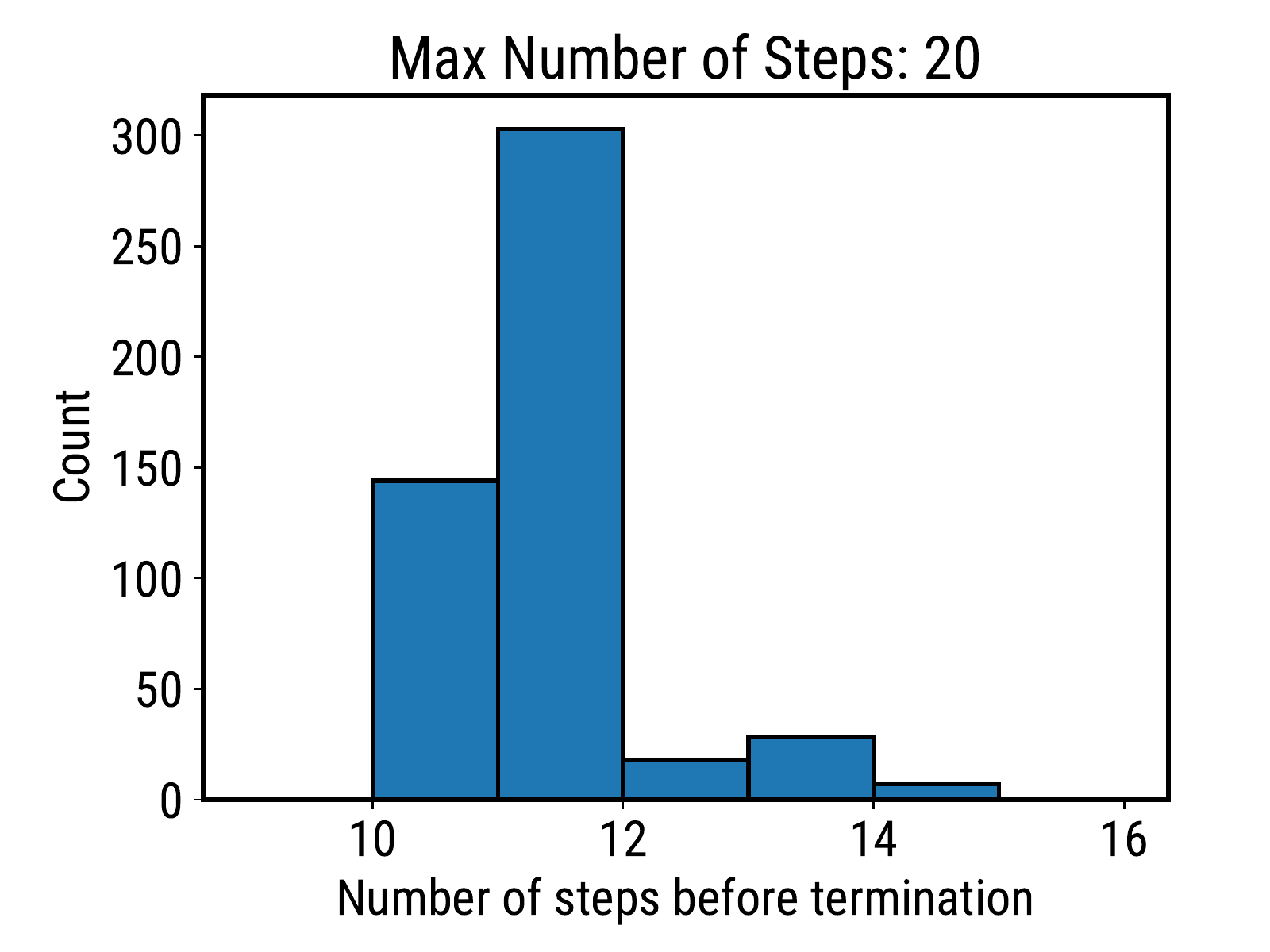} &    \includegraphics[width=0.5\textwidth]{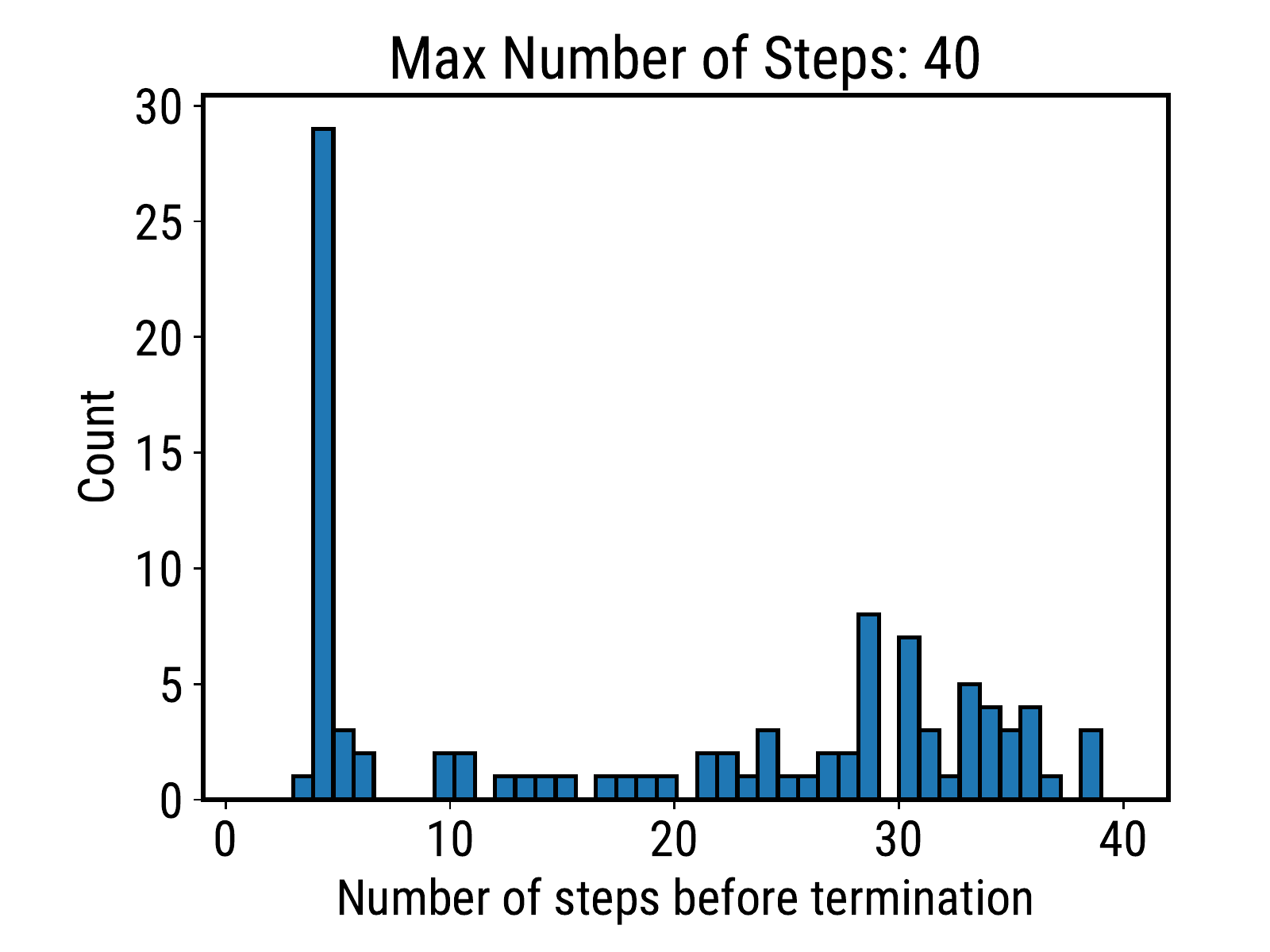} 
  \end{tabular}

  \caption{Histogram of number of steps before the policy
  chooses to stay at the same step (the ``no modification'' action, all subsequent actions are ``no modification"). (a)The task is to find a molecule whose molecular weight lies between 150 and 200. (b) The task is to find
  a molecule that maximizes the QED. Bootstrap is turned off in this experiment.}
  \label{fg:epi_len}
\end{figure}

\begin{figure}
    \centering
    \includegraphics{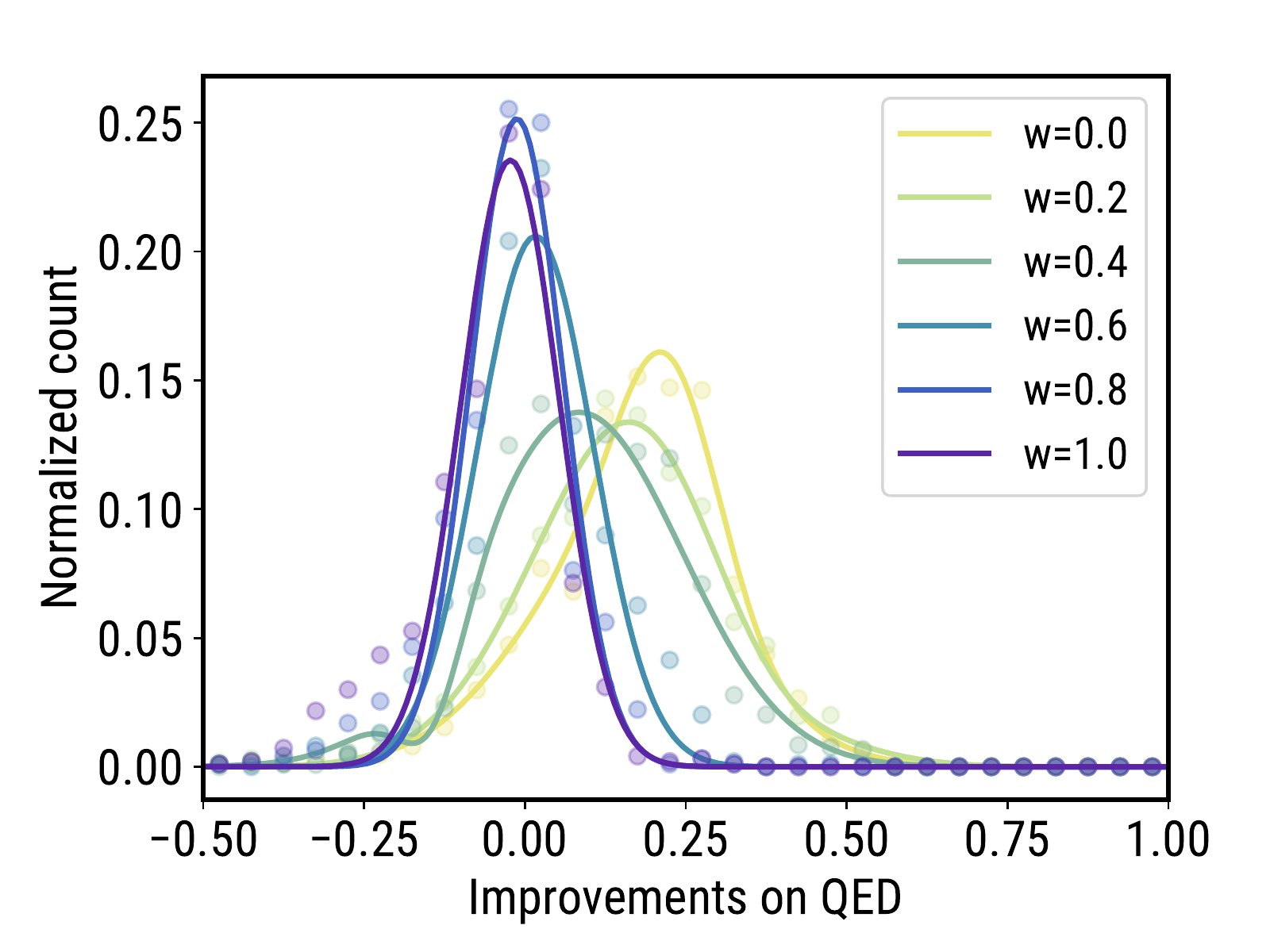}
    \caption{The empirical 
  distribution of the QED improvements in
  20 multi-objective optimization tasks.
  The variable $w$ in legends denotes the weight of the similarity in the multi-objective reward, while the QED score is weighted by ${(1-w)}$, i.e.
$ r = w \times \mbox{SIM}(s) + (1 - w) \times \QED(s)$.}
    \label{fig:qed_abs_imp}
\end{figure}

\begin{figure}
    \centering
    \includegraphics{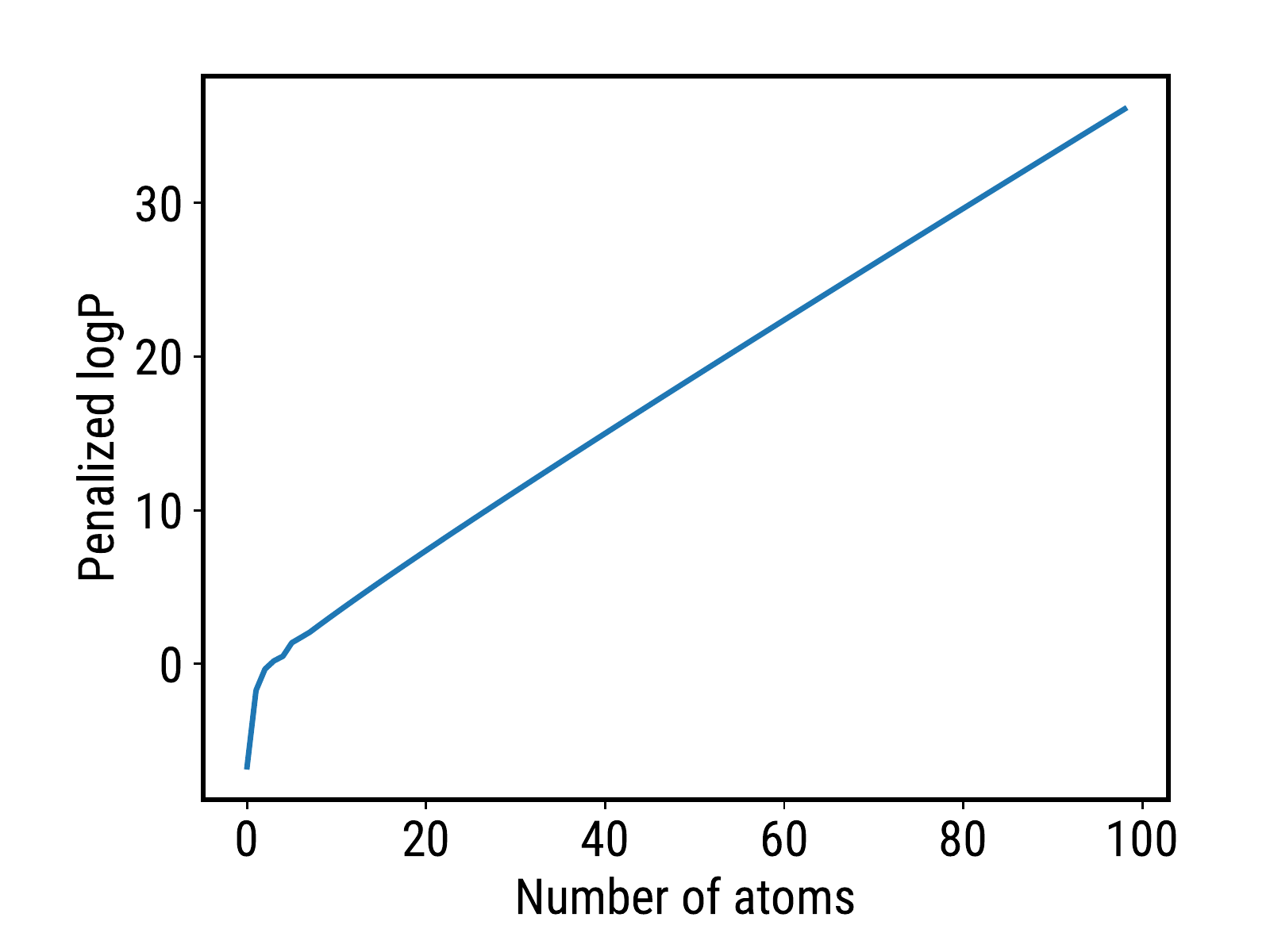}
    \caption{Penalized logP values of acyclic saturated alkane with different number of carbon atoms.}
    \label{fig:logp_vs_natoms}
\end{figure}

\clearpage

\section{Property Targeting}
\subsection{Single Property Targeting}
It is crucial to find molecules with properties close to a given
target in molecule design. Molecular weight and 
hydrophobicity are two important properties in drug design related to drug absorption. Therefore, we chose a target 
range of molecular weight (MW) and octanol-water partition coefficient (logP),
and measured the percentage of the molecules belonging to the
specified range. Given a target range of $[l, u]$, where $l$ is the lower bound
and $u$ is the upper bound, the reward function of a molecule $m$
is designed as follows:
\begin{equation}
    R(s) = \begin{cases}
		1 								& \mbox{if}\hspace{1em} p(m) \in [l, u] \\
		-\min\{|p(m) - l|, |p(m) - u|\} & \mbox{otherwise}
	\end{cases}
	\label{eq: reward_prop_tar}
\end{equation}

where $p(m)$ is the property value of molecule $m$. Intuitively, this reward is measuring 
the distance between the current property value
and the range we define. When the property value falls into the range we want,
a positive reward is given indicating the goal was reached. Otherwise, 
the reward is the negative minimum distance the property has to ``move'' 
in order to be in the desired range. The model was trained using the reward defined
in Eq. (\ref{eq: reward_prop_tar}), with the initial molecule being empty.
Evaluation was run for 500 episodes, and
the properties of the 500 generated molecules are reported.

\begin{table}
  \caption{The success rate of the property targeting task.}
  \label{tb:prop_tar}
  \begin{tabular}{ccccc}
    \hline
           & -2.5 $\leq$ logP $\leq$ -2 & 5 $\leq$ logP $\leq$ 5.5 & 150 $\leq$ Mw $\leq$ 200 & 500 $\leq$ Mw $\leq$ 550 \\
    \hline
    random walk\textsuperscript{\emph{a}} & 0.0\% & 1.0\% & 0.0\% & 0.0\% \\
    JT-VAE\textsuperscript{\emph{b}} & 11.3\% & 7.6\%  & 0.7\%   & 16.0\%  \\
    ORGAN\textsuperscript{\emph{b}}  & 0.0\%  & 0.2\%  & 15.1\%  & 0.1\%  \\
    GCPN\textsuperscript{\emph{b}}   & 85.5\% & 54.7\% & 76.1\%  & 74.1\%  \\
    ours   & \textbf{100\%}  & \textbf{100\%}  & \textbf{100\%}   & \textbf{100\%}  \\
    \hline

  \end{tabular}
  
  \raggedright
      \vspace{0.1in}
      {\small
      \textsuperscript{\emph{a}} ``random walk'' is a baseline that chooses a random action for each step.
     
     \textsuperscript{\emph{b}} values are reported in \citet{you2018graph}.
     }
\end{table}

 Using the same ranges chosen in 
\citet{you2018graph}, the effectiveness of the property 
targeted molecule optimization
is shown in Table~\ref{tb:prop_tar}. Our model outperforms 
others by reaching 100\% success
rates on all tasks. As stated in \citet{you2018graph}, the ranges are chosen such that
few molecules in ZINC dataset\cite{irwin2012zinc} are within that range.
Compared with other models, our model do not use expert 
pretraining on ZINC dataset, therefore the properties of the 
molecules generated is not limited 
by the properties of molecules in ZINC.

Note that since our deterministic policy only leads to 1--3 unique molecules, the success
rates in Table~\ref{tb:prop_tar} are not comparable. However, this experiment shows that MolDQN is able
to find molecules with arbitrary property values.

\subsection{Multi-Objective Property Targeting}

Here we want to illustrate that our model can find optimal molecules that
satisfy two constraints at the same time. Similar to the experimental
setup in \citet{li2018multi}, the objective is to find
optimal molecules which are close to specific 
Synthetic Accessibility (SA)\cite{ertl2009estimation} scores 
and Quantitative Estimate of Druglikeness 
(QED)\cite{bickerton2012quantifying}  values. Here four 
different targets are specified as follows: $c_1=(2.2, 0.84)$,
$c_2=(2.5, 0.27)$, $c_3=(3.8, 0.84)$, and $c_4=(4.8, 0.27)$, where
the first value is the target SA score and the second is the QED score.

The model was trained using the reward defined
below, starting with an empty initial molecule:
\[
R(s) = -\left(\left|\QED(m) - \QED_{\mbox{target}}\right| +  
       \left|\mbox{SA}(m) - \mbox{SA}_{\mbox{target}}\right| \right)
\]

\begin{table}
  \caption{Property statistics of the optimized molecules.}
  \label{tb:multi_obj_gen}
  \begin{tabular}{ccccccccc}
    \hline
    	   & \multicolumn{2}{c}{Target 1} & \multicolumn{2}{c}{Target 2}
    	   & \multicolumn{2}{c}{Target 3} & \multicolumn{2}{c}{Target 4}\\ 
           &  SAS  &  QED  &  SAS  &  QED &  SAS  &  QED &  SAS  &  QED \\
    \hline
    target value
    	& 2.200 & 0.840 & 2.500 & 0.270 & 3.800 & 0.840 & 4.800 & 0.270 \\
    mean \textsuperscript{\emph{a}}
    	& 2.303 & 0.859  & 2.564  & 0.251  & 3.806 & 0.834 & 4.799 & 0.272 \\
    standard deviation  \textsuperscript{\emph{a}}
    	& 0.109  & 0.012  & 0.114  & 0.009  & 0.074 & 0.012 & 0.069 & 0.005 \\
    mean absolute difference \textsuperscript{\emph{a}}
    	& 0.103  & 0.019  & 0.075  & 0.020  & 0.013 & 0.012 & 0.009 & 0.003 \\

    \hline
  \end{tabular}
  
  \raggedright
    \vspace{0.1in}
    {\small
      \textsuperscript{\emph{a}} Mean, standard deviation, and mean absolute
      difference denotes the statistics on unique generated molecules. 
      }
\end{table}

The distributions of SA score and QED for the molecules generated by our
model are shown 
in Table~\ref{tb:multi_obj_gen}. Even though SA scores and QED may change significantly with small modifications of the molecule, the properties of the generated molecules have a narrow distribution. 
These results illustrate that
explicit rewards on target values can lead to accurate targeted optimization with reinforcement learning.

\bibliography{references}